\documentclass[12pt]{iopart}

\usepackage{iopams} 
\usepackage[symbol]{footmisc}
\usepackage{cite}
\usepackage{slashbox}

\usepackage[utf8]{inputenc} % allow utf-8 input
\usepackage[T1]{fontenc}    % use 8-bit T1 fonts
\usepackage{hyperref}       % hyperlinks
\usepackage{url}            % simple URL typesetting
\usepackage{booktabs}       % professional-quality tables
\usepackage{nicefrac}       % compact symbols for 1/2, etc.
\usepackage{microtype}      % microtypography
\usepackage{enumitem}
\usepackage{amsmath}
\usepackage{wrapfig}
\usepackage{graphicx}
\usepackage{cancel}
\usepackage{soul}
\usepackage[dvipsnames]{xcolor}
\newcommand{\CellWithForcedBreak}[2][c]{\begin{tabular}[#1]{@{}c@{}}#2\end{tabular}}
\newcommand{\cL}{\mathcal{L}}
\newcommand{\cF}{\mathcal{F}}
\newcommand{\cN}{\mathcal{N}}
\newcommand{\cS}{\mathcal{S}}

\def\bea{\begin{eqnarray}}
\def\eea{\end{eqnarray}}

\def\be{\begin{equation}}
\def\ee{\end{equation}}

\def\eqn#1{eq.~\eqref{#1}}

\makeatletter
\DeclareRobustCommand*{\bfseries}{%
  \not@math@alphabet\bfseries\mathbf
  \fontseries\bfdefault\selectfont
  \boldmath
}

\makeatother

\usepackage{todonotes}

\begin{document}

\hfill SLAC-PUB-17774

\title[Transforming the Bootstrap]{Transforming the Bootstrap: Using Transformers to Compute Scattering Amplitudes in Planar $\mathcal{N}=4$ Super Yang-Mills Theory}

\author{Tianji Cai$^{a\setcounter{footnote}{0}\footnotemark\footnotemark\setcounter{footnote}{0}}$, Garrett W.\ Merz$^{b\footnotemark\footnotemark\setcounter{footnote}{0}}$, François Charton$^{c\footnotemark\setcounter{footnote}{0}}$, Niklas Nolte$^c$, Matthias Wilhelm$^d$, Kyle Cranmer$^b$, Lance J.\ Dixon$^a$}
\address{$^a$ SLAC National Accelerator Laboratory}
\address{$^b$ Data Science Institute, University of Wisconsin-Madison}
\address{$^c$ FAIR, Meta}
\address{$^d$ Niels Bohr Institute, University of Copenhagen}

\def\thefootnote{*}\stepcounter{footnote}\footnotetext{Denotes equal contribution}\stepcounter{footnote}\footnotetext{Authors to whom correspondence should be addressed}\def\thefootnote{\arabic{footnote}}

\vspace{10pt}
\begin{indented}
\item[] tianji@slac.stanford.edu, garrett.merz@wisc.edu, fcharton@meta.com, nolte@meta.com, matthias.wilhelm@nbi.ku.dk, kyle.cranmer@wisc.edu, lance@slac.stanford.edu 
\item[]Sept 2024
\end{indented}
\begin{abstract}
We pursue the use of deep learning methods to improve state-of-the-art computations in theoretical high-energy physics. Planar $\mathcal{N}=4$ Super Yang-Mills theory is a close cousin to the theory that describes Higgs boson production at the Large Hadron Collider; its scattering amplitudes are large mathematical expressions containing integer coefficients.  In this paper, we apply Transformers to predict these coefficients. The problem can be formulated in a language-like representation amenable to standard cross-entropy training objectives. We design two related experiments and show that the model achieves high accuracy ($>98\%)$ on both tasks. Our work shows that Transformers can be applied successfully to problems in theoretical physics that require exact solutions.  
\end{abstract}

%%%%%%%%%%%%%%%%%%%%%%%%%%%%%%%%%%%%%%%%%%%%%%%%%%%%%%%%%%%%%%%%%%%%%%%%%%%%%%%
\section{Introduction}

Particle physics at the energy frontier is entering an exciting new era of high-precision experiments, ushered in by the high-luminosity upgrade of the Large Hadron Collider (LHC). Exploiting the full physics potential of the experimental data requires substantial improvements in the predictions of Standard Model (SM)~\cite{Huss:2022ful} processes, both as backgrounds to new physics, and for measuring Higgs boson couplings and other SM parameters.

\begin{figure}
\centering
\begin{tabular}{cc}
\includegraphics[width=100mm]{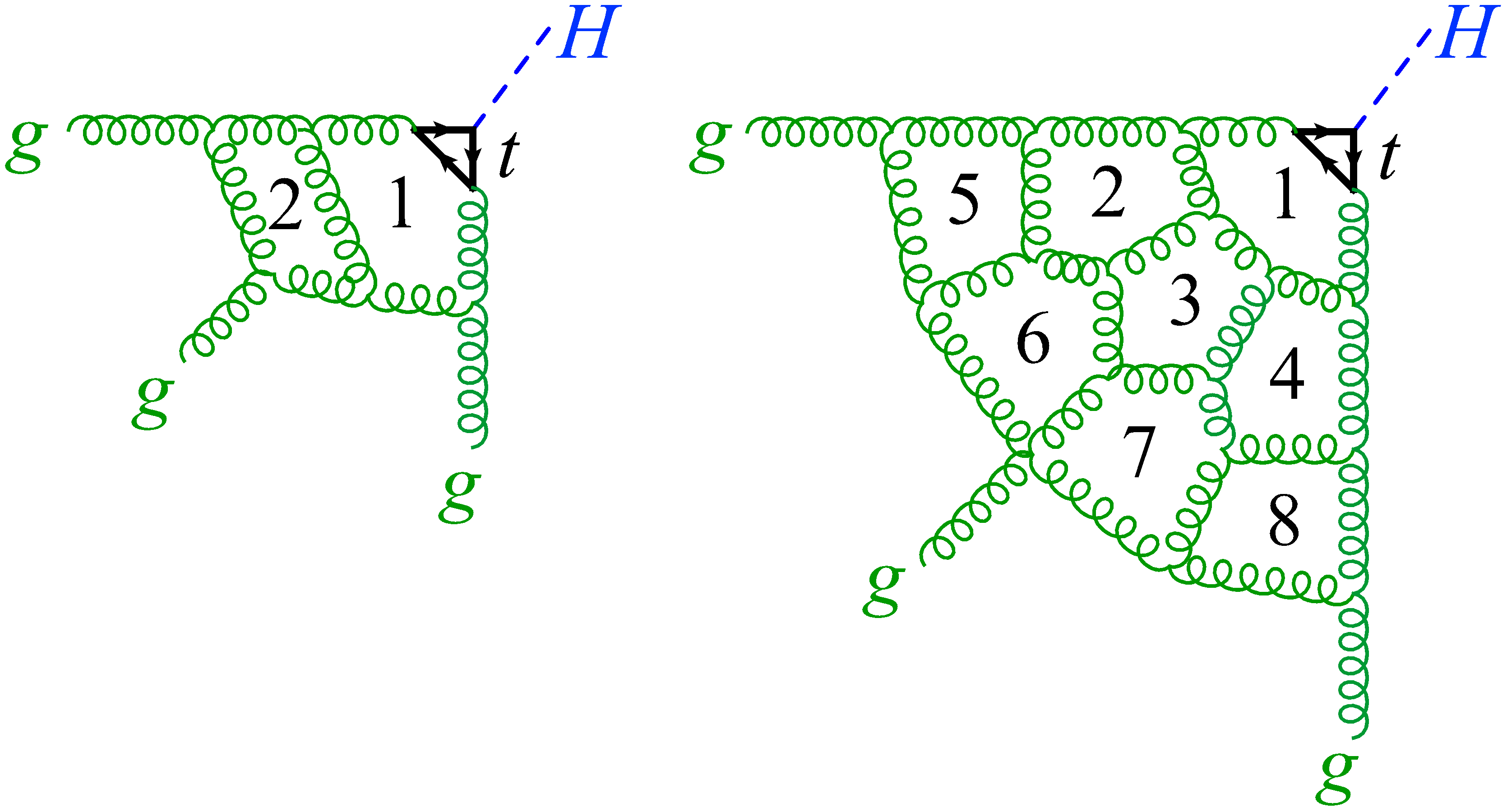} 
\end{tabular}
\centering
\caption{\small Sample Feynman diagrams for the process $gg \rightarrow Hg$ at two loops (left) and eight loops (right) in QCD. The same diagrams contribute in SYM, where the Higgs boson $H$ and top quark ($t$) triangle is replaced by a particular local operator in the theory, and the process is referred to as a form factor.}
\label{fig:feynmandiagrams}
\end{figure}

Many ingredients are necessary for these predictions, see e.g.~\cite{Heinrich:2020ybq} for a review. At the heart of all such calculations are scattering amplitudes---the fundamental quantum-mechanical building blocks for transition probabilities between asymptotic states. The conventional way to compute scattering amplitudes uses Feynman diagrams (see  Figure~\ref{fig:feynmandiagrams} for examples), which graphically organize a series of terms in a perturbative expansion. Performing high-precision calculations in the theory of quantum chromodynamics (QCD) requires Feynman diagrams containing at least two loops~\cite{Anastasiou:2015vya,Anastasiou:2016cez,Mistlberger:2018etf,Duhr:2019kwi,Dreyer:2016oyx,Duhr:2020seh,Duhr:2020sdp,Duhr:2021vwj}.
Each loop represents intermediate-state virtual particles whose unobserved momenta must be integrated over. Each successive order of precision demands the addition of another loop to the diagram.  Unfortunately, the number of possible Feynman diagrams, and thus the number of integrals that must be performed, grows factorially with loop order, quickly making these calculations intractable.

A recently-developed alternative technique, known as the {\it amplitude bootstrap}~\cite{Dixon:2011pw, Caron-Huot:2016owq,Caron-Huot:2019vjl}, attempts to directly construct candidate solutions for multi-loop amplitudes. 
It has mainly been applied so far to a simpler relative of QCD, called {\it planar} ${\cal N}=4$ {\it super-Yang-Mills theory} (SYM).
The amplitude bootstrap circumvents many of the computational and numerical challenges that arise from the Feynman diagram approach. It leverages the rich, yet highly constrained, analytical structure of amplitudes that arises from the particular recurrent features of the integrals involved. Using this technique, the form of the amplitude can be determined \emph{a priori}, and the finite-dimensional solution space at a given loop order can be strongly constrained through a large system of linear relations with integer coefficients. Many of the linear relations are found by analyzing the lower-loop results.
A small number of additional constraints, derived from behavior in physical limits, can then be applied in order to obtain a unique solution.

The amplitude bootstrap allows for the computation of amplitudes up to eight loops in SYM~\cite{Dixon:2022rse}, \emph{vs.}~two loops using traditional Feynman diagram methods for the same quantity in QCD \cite{Gehrmann:2011aa}, as depicted in Figure~\ref{fig:feynmandiagrams}. However, when using the bootstrap technique, the number of linear relations to be solved and the number of unknown coefficients both increase by a factor of about 4 at each subsequent loop order~\cite{Dixon:2022rse}. 
Since it is not possible to determine the minimal set of independent equations in advance, the number of relations can be several times larger than the number of unknown coefficients. The computational cost of generating and solving the equations thus increases by at least a factor of $4^2 = 16$ at each loop order. Performing computations by this method becomes infeasible beyond about loop $L=9$. This situation necessitates the development of new methods that may exploit hitherto unobserved patterns in the data in order to simplify the computation of scattering amplitudes at higher loops.

Notably, amplitudes in SYM can be expressed as sets of tokenizable ``words'' with integer coefficients; see Section~\ref{sec:theory}. We refer to these words as \textit{keys} which index into the coefficients, and to a key-coefficient pair as an \textit{element}. Exploiting the many linear relations between these elements amounts to solving a large system of linear equations (as explained above),  
where the integer solutions are hard to discover, but easy to verify. 

In our case, verification includes first checking that the proposed expression is a legitimate function in the appropriate space (see~\ref{app:symb_map}), and then comparing its behavior in a physical limit to the predictions of the form-factor operator product expansion (FFOPE)~\cite{Basso:2013vsa,Sever:2020jjx}. The latter constraint is stringent enough to fix the answer uniquely~\cite{Dixon:2022rse}.

Since both the keys and the integers can be represented as sequences of tokens, we can train deep learning models such as Transformers to predict the coefficient associated with each key.

Transformers \cite{Vaswani:2017lxt} are incredibly versatile neural network architectures that employ an {\it attention mechanism} \cite{Bahdanau:2014ghw} to learn complex nonlinear relationships between input features. They have revolutionized fields ranging from natural language processing \cite{BERT} and computer vision \cite{ViT} to formal symbolic mathematics \cite{FunSearch, charton2022linear}. Inspired by these many recent successes, we apply Transformers to two sets of experiments; see Section~\ref{sec:settings} for our model setup.

We first show in Section~\ref{sec:coefffromword} that Transformers can accurately predict elements of the solution at a given loop order when trained on other elements at the same loop. In Section~\ref{sec:quads}, we show that the model is able to do this even when data is presented in a highly compressed format. To make these predictions successfully, many features of the complex relationships between individual terms must be learned by the model. In the subsequent Section~\ref{sec:relations}, we explore how a number of the known linear relations are learned as a function of training epoch, and use this information to draw conclusions about the learning dynamics of the model. In Section~\ref{sec:mixed_loop}, we show that augmenting a small amount of training data at a given loop with data from a lower loop improves performance, and discuss prospects for future multi-loop experiments.

The mathematical structure of the problem hints that some elements of the solution at higher loops may be determined using related elements of the solution at lower loops. Our second goal is therefore to discover this relationship implicitly, assuming it exists. In Section~\ref{sec:strikeout}, we train Transformers to predict coefficients of terms at loop $L+1$, given a set of coefficients of potentially related terms at loop $L$. We also perform a number of ablations to determine conditions under which this relationship is no longer learnable. This study allows us to uncover certain features of the cross-loop relationship, which may prove crucial in further developing the bootstrap program. 

We emphasize that the goal of these experiments is not to optimize performance.  Rather it is to show that properties of amplitudes can be learned by AI models, and to probe the learning dynamics.

Additionally, this paper contains several Appendices. In \ref{app:symb_map}, we further describe the mathematical formalism by which scattering amplitudes are expressible in a language-like fashion, via their {\it symbols}, which are sums of pairs of keys and integer coefficients. In \ref{app:rels_list}, we give a more comprehensive list of linear relations between symbol terms and evaluate them as a function of training epoch. In \ref{app:ablations}, we perform a number of architecture ablations and evaluate their effects on the coefficient prediction task of Section~\ref{sec:coefffromword}. In \ref{app:recurrence}, we perform additional ablation experiments in the manner of Section~\ref{sec:strikeout} in order to further characterize the correspondence between elements at different loop orders.

\subsection{Related Work}

The amplitude bootstrap program has a long tradition, dating back to ref. ~\cite{Dixon:2011pw}. A recent review can be found in ref. ~\cite{Caron-Huot:2020bkp}, and the specific data we use is from ref. ~\cite{Dixon:2022rse}. Our work supplements the traditional approach by offering a novel problem-solving framework, where human intelligence is augmented by artificial intelligence to further push the state-of-the-art for amplitude calculations.

In a similar spirit, a number of recent works have also leveraged deep learning to tackle \textit{analytical} calculations in theoretical physics. In particular, a sequence-to-sequence Transformer has been used to compute the squared amplitude of a particle interaction symbolically~\cite{Alnuqaydan:2022ncd}; deep reinforcement learning has been applied to explore the landscape of string vacua~\cite{Halverson:2019tkf}; and Transformers have been employed to simplify polylogarithms~\cite{Dersy:2022bym}, which are complicated mathematical functions entering multi-loop amplitudes similar to those we study (see Section~\ref{ssec:symbols}). However, no previous work has used Transformers to perform computations in the amplitude bootstrap paradigm.

Methodologically, our work is closely related to recent works using Transformers for \textit{symbolic mathematical data}. For example, Transformers have been taught to perform mathematical tasks such as solving differential equations~\cite{lample2019deep}, learning recurrent sequences~\cite{dascoli2022deep}, and finding the greatest common divisor of number pairs ~\cite{charton2024learning}.
A comparable approach has also been used to solve linear algebra tasks~\cite{charton2022linear}, including eigenvector decomposition and matrix inversion, which share many structural similarities with our amplitude bootstrap method. Additionally, our first experiment, in which we use some elements of a scattering amplitude to predict others, can be framed as a tensor completion problem---at a fixed loop order, the model must learn to fill in unseen elements of the solution based on elements it has seen at training time. This task is similar in some respects to low-rank matrix completion~\cite{DBLP:journals/corr/abs-1211-4116}, where Transformers have previously been employed~\cite{lee2023teaching}.

%%%%%%%%%%%%%%%%%%%%%%%%%%%%%%%%%%%%%%%%%%%%%%%%%%%%%%%%%%%%%%%%%%%%%%%%%%%%%%%
\section{Three-Gluon Form Factors in Planar \texorpdfstring{$\cN=4$}{N=4} SYM Theory}
\label{sec:theory}
The amplitude bootstrap program has seen substantial success in planar $\cN=4$ super-Yang-Mills theory~\cite{Brink:1976bc}. Similar to QCD, SYM contains gluons which self-interact; but instead of including quarks, it contains four gluinos and six scalars. The gluons, gluinos and scalars are all massless, and all transform into each other under the $\cN=4$ supersymmetry. They all have the same number of internal ``color'' degrees of freedom.  We take the number of colors $N_c$ to infinity, and refer to the Feynman diagrams that contribute to the scattering of these massless particles as \textit{planar}.
As a theoretical laboratory or model system for QCD, SYM allows us to see much further into the perturbative expansion than QCD. For example, a class of SYM amplitudes was recently computed to {\it eight} loops~\cite{Dixon:2020bbt,Dixon:2022rse}. These amplitudes, referred to as \textit{three-gluon form factors $\cF_{\tt 3gFF}$}, involve three massless gluons and a massive color-singlet operator. The operator couples to gluons very similarly to how the Higgs boson does in the limit of a very heavy top quark. Thus $\cF_{\tt 3gFF}$ is the SYM analog of the QCD process $gg\to Hg$, which is known only to two loops~\cite{Gehrmann:2011aa}. In fact, part of the QCD form-factor result (the so-called ``highest-weight'' part) is identical to the SYM result~\cite{Brandhuber:2012vm,Duhr:2012fh}. Figure~\ref{fig:feynmandiagrams} shows sample Feynman diagrams for this process to the current highest calculable loop order in QCD and SYM, respectively.

In this work, we focus exclusively on the three-gluon form factors $\cF_{\tt 3gFF}$ in planar $\cN=4$ SYM.  The known results up to loop $L=7$ are used for model training and evaluation.

\subsection{Symbols: A Simple Language for Amplitudes}
\label{ssec:symbols}

The three-gluon form factors, like many other amplitudes in SYM, can be expressed in terms of functions called {\it generalized polylogarithms}. They are multiple, iterated integrations of rational functions. In SYM, the calculation of scattering amplitudes at loop order $L$ requires integrals that are iterated $2L$ times. 
Such amplitudes $\cF^{(L)}$ can be characterized by another mathematical object known as the \textit{symbol}~\cite{Goncharov:2010jf}:
\be
\cS[{\cF}^{(L)}] =
       \sum_{l_{i_1},\ldots,l_{i_{2L}}\in\cL_m} C^{l_{i_1},\ldots, l_{i_{2L}}} \,
l_{i_1} \otimes \cdots \otimes l_{i_{2L}} \,.
\label{symbF}
\ee
Here $\cL_m = \{l_1, \dots, l_{m}\}$ is the {\it symbol alphabet} containing $m$ {\it letters} $l_i$, which are in turn functions of the particles' four-momenta, and $C^{l_{i_1},\ldots, l_{i_{2L}}}$ is a ${2L}$-fold tensor of integer coefficients, most of which are zero. In other words, a solution for the symbol at loop $L$ can be represented by $m^{2L}$ integers, with each sequence of $2L$ letters (i.e., $l_{i_1} \otimes \cdots \otimes l_{i_{2L}}$) serving as a key indexing into the integer-valued tensor $C^{l_{i_1},\ldots, l_{i_{2L}}}$. Symbol terms encode information about the derivatives and discontinuities of the polylogarithms, and do not correspond directly to individual Feynman diagrams. More details about the map from generalized polylogarithms to symbols are given in~\ref{app:symb_map}.

The alphabet of $\cF_{\tt 3gFF}$ is one of the simplest among all amplitudes and contains only six letters, i.e., $m=6$:
\be
\cL_{\tt 3gFF} = \{a,b,c,d,e,f\} \,,
\label{F3alphabet}
\ee
cf.\ \ref{app:symb_map}. 
These letters are Lorentz-invariant functions of the gluons' four-momenta. 
Via their definition they transform under a dihedral symmetry with two generators:
\be
\text{\textbf{cycle:} } \{a, b, c, d, e, f\} \to \{b, c, a, e, f, d\} \
\text{, and \textbf{flip:} } \{a, b, c, d, e, f\} \to \{b, a, c, e, d, f\} \,.
\label{dihedralsymm}
\ee 
The $L$-loop form factor $\cF_{\tt 3gFF}^{(L)}$ is invariant under dihedral transformations for any $L$.

As concrete examples, the symbols for $\cF_{\tt 3gFF}^{(L)}$ at loops $L=1$ and $L=2$ contain only 6 and 12 nonvanishing terms, respectively:
\bea
\cS[\cF_{\tt 3gFF}^{(1)}] &=& (-2) \Bigl[ b \otimes d + c \otimes e + a \otimes f
  + b \otimes f + c \otimes d + a \otimes e \Bigr] \,, \nonumber\\
\cS[\cF_{\tt 3gFF}^{(2)}] &=& 8 \Bigl[
    b \otimes d \otimes d \otimes d
  + c \otimes e \otimes e \otimes e
  + a \otimes f \otimes f \otimes f \nonumber\\
&&\hskip0.6cm\null
  + b \otimes f \otimes f \otimes f
  + c \otimes d \otimes d \otimes d
  + a \otimes e \otimes e \otimes e \Bigr] \nonumber\\
&&\hskip0cm\null
+ 16 \Bigl[
    b \otimes b \otimes b \otimes d
  + c \otimes c \otimes c \otimes e
  + a \otimes a \otimes a \otimes f \nonumber\\
&&\hskip0.9cm\null
  + b \otimes b \otimes b \otimes f
  + c \otimes c \otimes c \otimes d
  + a \otimes a \otimes a \otimes e \Bigr] \,.
\label{symbF3onetwoloop}
\eea

Usually, we omit the tensor product ``$\otimes$'' and use for example ``$\tt{bd}$'' as shorthand for ``$b \otimes d$'', calling it a \emph{word} or a \emph{key}.  For $L=1$, the key $\tt{bd}$ then indexes into the tensor $C^{l_1,l_2} = C^{b, d}$, mapping onto the integer coefficient $-2$. As another example, the key $\tt{ab}$ never appears in $\cS[\cF_{\tt 3gFF}^{(1)}]$, and therefore $\tt{ab}$ maps onto a coefficient of $0$.  The invariance of the form factor under dihedral transformations (\ref{dihedralsymm}) relates all terms with the same coefficients in \eqn{symbF3onetwoloop} to one another.  Hence at $L=1$ ($L=2$) there are really only 1 (2) nonzero terms to predict.

One can recover the iterated integral representation from the symbol, up to constants that can be recovered in a similar fashion. For the one-loop form factor given in eq.~(\ref{symbF3onetwoloop}), its iterated integral representation is
\begin{equation}
\cF_{\tt 3gFF}^{(1)} = 2 \, \big[ {\rm Li}_2(1-bc) + {\rm Li}_2(1-ca) + {\rm Li}_2(1-ab) \big]\,, 
\end{equation}
where ${\rm Li}_2(x) = -\int_0^x dt \ln(1-t)/t$. 
The iterated integral representation of the two-loop form factor $\cS[\cF_{\tt 3gFF}^{(2)}]$ (in a different normalization) is given in eq.~(4.32) of ref.~\cite{Brandhuber:2012vm}.

In general, $\cS[\cF_{\tt 3gFF}^{(L)}]$ is a sum of $6^{2L}$ \emph{elements} (i.e., the monomials in \eqn{symbF3onetwoloop}) containing a key which is a $2L$-sequence of letters, and an associated integer coefficient.
This large number of $6^{2L}$ elements can be substantially reduced via a set of conditions on the symbol that restrict which letters can appear next to each other. Therefore, most of the $6^{2L}$ coefficients are zero, and most zeros can be accounted for by the following two simple rules: 
\begin{itemize}[nosep]
\item \textbf{adjacency rule:} any key including one of the subsequences $\tt{ad}$, $\tt{da}$, $\tt{de}$ (or their dihedral images) has zero coefficient~\cite{Dixon:2022rse}.
\item \textbf{prefix/suffix rule:} any key beginning with $\tt{d}$, $\tt{e}$ or $\tt{f}$ or ending with $\tt{a}$, $\tt{b}$ or $\tt{c}$ has zero coefficient.
\end{itemize}
We call such zero coefficients {\it trivial zeros}. Table~\ref{tab:elem_count} records the actual numbers of nonzero coefficients in the symbol at different loop orders, as well as the naive $6^{2L}$ counts and the counts excluding trivial zeros. The table shows that, at loop $L=6$ and higher, about half of the terms allowed by the above two simple rules still have a zero coefficient.  We refer to such terms as {\it nontrivial zeros}.

\begin{table}[h]
    \small
    \centering
    \begin{tabular}{lccccccccc}
        \toprule
        Loop & $1$ & $2$ & $3$ & $4$ & $5$ & $6$ & $7$ & $8$  \\
        \midrule
        Total ($6^{2L}$) & $36$ & $1,296$ & $46,656$ & $1.7\cdot 10^6$ & $6.0\cdot 10^7$ & $2.2 \cdot 10^9$ & $7.8\cdot 10^{10}$ & $2.8\cdot 10^{12}$\\
        W/O trivial zeros & $6$ & $102$ & $1,830$ & $32,838$ & $589,254$ & $1.1\cdot 10^7$ & $1.9\cdot 10^8$ & $3.4\cdot 10^9 $\\
        Total nonzero & $6$ & $12$ & $636$ & $11,208$ & $263,880$ & $4.9\cdot 10^6$ & $9.3\cdot 10^7$ & $1.7\cdot 10^9$\\
       \bottomrule
    \end{tabular}
    \caption{\small Elements in the symbol $\cS[\cF_{\tt 3gFF}^{(L)}]$ for loops $L=1$ to 8. }
    \label{tab:elem_count}
\end{table}

\begin{figure}[h]
\center
\begin{tabular}{cc}
\includegraphics[width=0.24\linewidth]{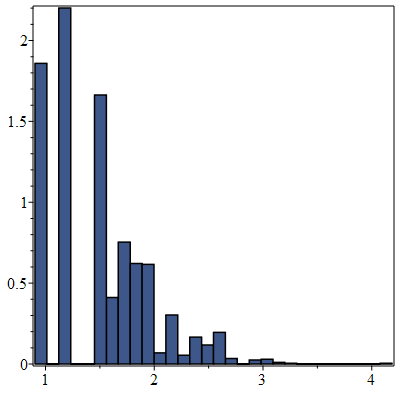}
\includegraphics[width=0.24\linewidth]{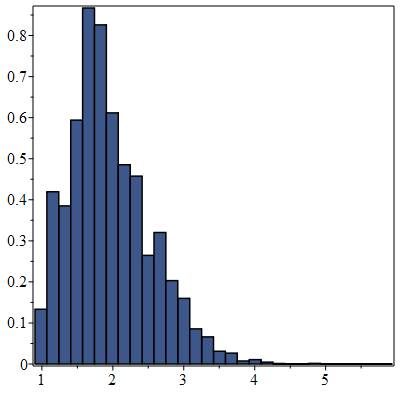}
\includegraphics[width=0.24\linewidth]{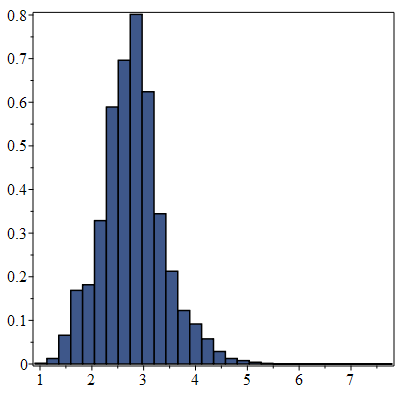}
\end{tabular}
\caption{Histograms of the symbol coefficients for the
  three-gluon form factor at 4, 5, and 6 loops.  The horizontal axis is the base 10 logarithm
  of the magnitude of the coefficient. The vertical axis is the (arbitrarily normalized) frequency with which coefficient magnitudes occur in the form factor.
  }
\label{fig:IIueq1Hist}
\end{figure}

Figure~\ref{fig:IIueq1Hist} shows the distribution of magnitudes of nonzero integer coefficients of $\cS[\cF_{\tt 3gFF}]$ at loops $L= 4, 5, 6$ on a log scale. Note that the magnitudes of the coefficients grow quickly from lower to higher loops, a domain shift which may pose a challenge when attempting to train a model to generalize across loop orders.

%%%%%%%%%%%%%%%%%%%%%%%%%%%%%%%%%%%%%%%%%

\subsection{Linear Relationships among Symbol Elements}
\label{ssec:reldef}

In addition to the trivial zeros defined above, there exist many linear correlations in the symbol which highly constrain the values of many coefficients. 
We study three types of linear relations in this work---the integrability relations, the multiple-final-entry relations, and the triple-adjacency relation.

Many of these constraints are inspired by empirical observations and have deep physical roots yet to be understood. However, some are based on rather elementary mathematical considerations. For example, one important constraint is {\it functional integrability}: a random multi-variate symbol is not the symbol of any function, because mixed partial derivatives must commute. This requirement 
correlates large sets of coefficients with specific adjacent letter pairs. One such integrability relation reads
\be
F^{a,b}+F^{a,c}-F^{b,a}-F^{c,a} = 0,
\label{integ1_main}
\ee
which correlates the coefficients of four terms in the symbol at a time. 
Here $F^{a,b}$ is the abbreviation of $C^{l_1,\dots, l_{i-1}, a, b, l_{i+2},\dots, l_{2L}}$, the coefficient corresponding to the key ``$\tt{\dots a b \dots}$'' where the letter pair $\tt{ab}$ can appear in any pair of adjacent positions (or \textit{slots}) in the key. The remaining letters (indicated by ``$\tt{\dots}$'') may take any values but must be the same for each of the terms in the relation. 
In other words, \eqn{integ1_main} can be written equivalently as
\begin{equation}
\begin{aligned}
&C^{l_1,\dots, l_{i-1}, a, b, l_{i+2},\dots, l_{2L}}
+C^{l_1,\dots, l_{i-1}, a, c, l_{i+2},\dots, l_{2L}}\\
&-C^{l_1,\dots, l_{i-1}, b, a, l_{i+2},\dots, l_{2L}}
-C^{l_1,\dots, l_{i-1}, c, a, l_{i+2},\dots, l_{2L}}=0,
\end{aligned}
\label{integ1_main_full}
\end{equation}
for any choice of $l_1,\ldots,l_{i-1},l_{i+2},\ldots,l_{2L} \in \cL_{\tt 3gFF}$,
and it remains valid for any position $i$ in the $2L$-length keys. 

A concrete instance of relation~(\ref{integ1_main_full}) at five loops is 
\be
C^{c,\underline{a,b},c,a,b,d,c,c,d} + C^{c,\underline{a,c},c,a,b,d,c,c,d} - C^{c,\underline{b,a},c,a,b,d,c,c,d} - C^{c,\underline{c,a},c,a,b,d,c,c,d} = 0,
\label{integ1_instance}
\ee
where the relevant adjacent letter pairs are underlined and the four integer coefficients are $72$, $-88$, $-72$, $56$, respectively. We thus have $72 + (-88) - (-72) - 56 = 0$, satisfying \eqn{integ1_instance}. Another integrability relation correlates 14 coefficients at a time, the longest linear relation currently known.

While the integrability and triple-adjacency relations occur in all adjacent pairs of slots in the key, the final-entry conditions relate only sets of keys that have the same beginnings, but different suffixes. 

In general, these linear relationships can serve as an excellent probe of the learning dynamics of the models, informing us about which properties of the symbol are learned at different training stages. We explore a subset of them in Section~\ref{sec:relations}, and give a more comprehensive discussion in \ref{app:rels_list}.

\subsection{Compact Symbol Representations}

With increasing loop order, the number of elements in the symbol becomes very large, as can be seen in Table~\ref{tab:elem_count}: at $8$ loops, the symbol has around $1.7$ billion terms. A compact symbol representation is therefore necessary at high loops. We derive one such compact representation by noticing that the number of independent suffixes for keys of nonzero coefficients is very limited.

Applying the multiple-final-entry relations (see~\ref{app:rels_list}) and dihedral symmetry,  
one notices that all terms in the symbol can be related to terms ending in the following $8$ sequences of four letters:
$\tt{dddd}, \tt{bbbd}, \tt{bdbd}, \tt{bbdd}, \tt{dbdd}, \tt{fbdd}, \tt{dbbd}$, and $\tt{cddd}$. 
We thus create a new {\it quad representation} where $8$ new tokens are added to represent these $8$ suffixes. All keys at a given loop order are then represented by their first $2L-4$ letters plus one of the eight quad suffix letters.

Represented in the quad format, there are only 391,570 keys for the loop $L=6$ symbol, in contrast to the 5 million keys in the original uncompressed format. 
Furthermore, compressing the data using the quad representation naturally removes the dihedral symmetry and all multiple-final-entry relations that involve only the last four entries. It is therefore interesting to see how the model will perform when it is presented with the more ``efficient'' quad format, versus having to learn these relations from the full symbol. 

An even more compact {\it octuple representation} can be achieved by considering the last $8$ letters in each element. There are $93$ possible final-entry octuples after factoring out dihedral symmetry. This representation gives 16,971 keys in the $L=6$ symbol, 312,463 in the $L=7$ symbol, and $5.6$ million in the $L=8$ symbol. In the current study, we do not use the octuple representation, since the highest loop order under consideration is $L=7$. However, the octuple representation may become necessary in the future to push past $L=8$, the highest loop order currently known.

%%%%%%%%%%%%%%%%%%%%%%%%%%%%%%%%%%%%%%%%%%%%%%%%%%%%%%%%%%%%%%%%%%%%%%%%%%%%%%%%%%%%%%%%%%%%%%%%%%%%%%%%%%%%
\section{Implementation Details}
\label{sec:settings}
\vspace{-.3em}

In this section, we briefly discuss the default architecture and tokenization scheme for the later experiments.

Due to the discrete nature of our problem, all tasks are framed as sequence-to-sequence translation problems: coefficients and keys are both encoded as sequences of tokens, and the model is trained to minimize the cross-entropy of the probability distribution for the predicted coefficient sequence with the ground-truth solution. At loop $L$, keys are encoded as sequences of $2L$ letter tokens, e.g., \texttt{`a, a, b, d, d, c, e, e'}.  While several recent works have explored different ways to tokenize integers \cite{golkar2023xval,Nogueira}, we simply encode coefficients as sequences of numerical tokens in base $1000$; e.g., $12334$ as \texttt{`+, 12, 334'}, with the sign first \cite{wengersalsa,dascoli2022deep}. To preserve syntax, zero coefficients are arbitrarily assigned a sign token of \texttt{`+'}.

In most experiments, we use encoder-decoder Transformers, which contain a bidirectional Transformer encoder and an autoregressive Transformer decoder linked by a cross-attention mechanism \cite{Vaswani:2017lxt}. Both encoder and decoder have the same number of layers (up to $8$), the same number of attention heads ($8$ or $16$), and the same dimension ($d=256$, $d=512$ or $d=1024$). For all models, the tokenizer dimension and Transformer dimension $d$ are the same (i.e., $256$, $512$, or $1024$). Henceforth, we describe a model with $N$ layers in the encoder and $N$ layers in the decoder as an {\it $N$-layer Transformer}.

Overall, our models have between $4.5$ and $245$ million trainable parameters, and the best performance on many of our experiments is obtained with models with fewer than $35$ million parameters. In contrast, many popular large language models \cite{GPT,touvron2023llama} have tens of billions of parameters. Following similar observations on AI-for-mathematics applications \cite{charton2022linear}, we are able to achieve very good results with these small Transformers trained on domain-specific data. In all experiments except where noted, we use the smallest model that can perform the task with large ($> 98\%$) accuracy, where accuracy is defined per element rather than per individual token. 

In the default model, we use a learnable positional encoding in both the encoder and decoder \cite{Vaswani:2017lxt}; alternative schemes are explored in \ref{app:ablations}. The optimizer is Adam \cite{kingma2017adam}, with a learning rate of $10^{-4}$ and a flat learning rate schedule. We do not observe significant performance improvements from changes to the learning rate schedule or addition of learning-rate warmup. We believe this is due to the small size of the Transformers we use: in our experience, warmup becomes more useful for larger models.

All models are implemented in PyTorch \cite{pytorch} and trained on a single NVIDIA V100 GPU with $32$ GB of memory, or on larger architectures (A100). 

Throughout the paper, we define an epoch as a pass over 300,000 key-coefficient pairs, as opposed to the more common definition of one full pass over all training data. This makes the notion of epoch size more comparable between different experiments that employ different amounts of data at different loops. At the end of each epoch, the model is evaluated on a held-out test set. The train-test split is performed randomly. In all experiments, metrics such as accuracy are quoted on the held-out test set, not the full symbol. For example, for a training set of 100,000 elements and a test set of 1,000 elements, an accuracy of 90\% refers to predicting 900 of those 1,000 elements correctly.

Accuracies are measured on a test set of $N_{test}$ randomly sampled elements. The accuracy for each element is a binary variable (correct $=1$, wrong $=0$), so we may assume that the errors are binomial. Applying the central limit theorem to the average, we have a 95\% confidence interval of $1.96 \times \sqrt{a(1-a)/N_{test}} \approx 2\sqrt{(1-a)/N_{test}}$ due to statistical error, for an accuracy $a\approx1$. For $a=99\%$ and a test set size of $N_{test}=10,000$, this interval is $0.2\%$. There is also uncertainty in the accuracy due to the model's initialization.

Because the goal of these experiments is not to optimize final accuracy, we do not perform any fine-grained hyperparameter scans. For this reason, we do not typically have a separate validation set in addition to the test set.  However, in order to determine whether the specific choice of training and test set might play a role, in~\ref{app:archscan} we repeat one of our experiments for a few different train/test splits using the same set of hyperparameters chosen for the main work. We find qualitatively similar performance.

%%%%%%%%%%%%%%%%%%%%%%%%%%%%%%%%%%%%%%%%%%%%%%%%%%%%%%%%%%%%%%%%%%%%%%%%%%%%%%%
\section{Predicting Symbol Coefficients from Keys}
\label{sec:coefffromword}
\vspace{-.3em}

In the first experiment, we train Transformers to predict coefficients from their keys. The models are trained on a fraction of the symbol at a given loop, $L=5$ or $6$, and are tasked to predict the remaining terms. Because most coefficients are zero, we split the problem into two separate tasks: predicting whether the coefficient corresponding to a given key is zero, and predicting a nonzero coefficient from the associated key.

We first train $1$-layer Transformers with dimension $d=256$ and $8$ attention heads (i.e., $4.5$ million parameters) to predict whether coefficients are zero or nonzero. We construct a dataset consisting of all nonzero-coefficient elements in the symbol plus an equal number of zero-coefficient elements. Here, zero-coefficient elements are selected randomly from the pool of all possible zeros, and the majority of zeros are thus trivial (as defined in Section \ref{ssec:symbols}). We explore an alternative prescription to handle the nontrivial zeros in Section~\ref{sec:relations}.

At $L=5$, the model correctly classifies $99.96\%$ of elements in a test set of 10,000 examples, after only one epoch (corresponding to observing only $57\%$ of the symbol). At $L=6$, the model correctly classifies $99.91\%$ of the test set after one epoch, i.e., after observing only $3\%$ of the symbol, and $99.97\%$ after two epochs (i.e., after observing $6\%$ of the symbol). Distinguishing nonzero coefficients from these mostly trivial zeros thus appears to be a very simple problem for Transformers.

Predicting nonzero coefficients from their keys proves to be more difficult, necessitating larger models. We train $2$-layer Transformers with dimension $d=512$ and $8$ attention heads, for loops $L=5$ and $L=6$. At $L=5$, models are trained on 163,880 nonzero-coefficient elements (i.e., $62\%$ of the symbol), and tested on 100,000 elements. At $L=6$, models are trained on 4,816,466 nonzero-coefficient elements (i.e., $98\%$ of the symbol), and are again tested on 100,000 elements. 

\begin{figure}
\centering
\begin{tabular}{cc}
    \includegraphics[width=0.5\linewidth]{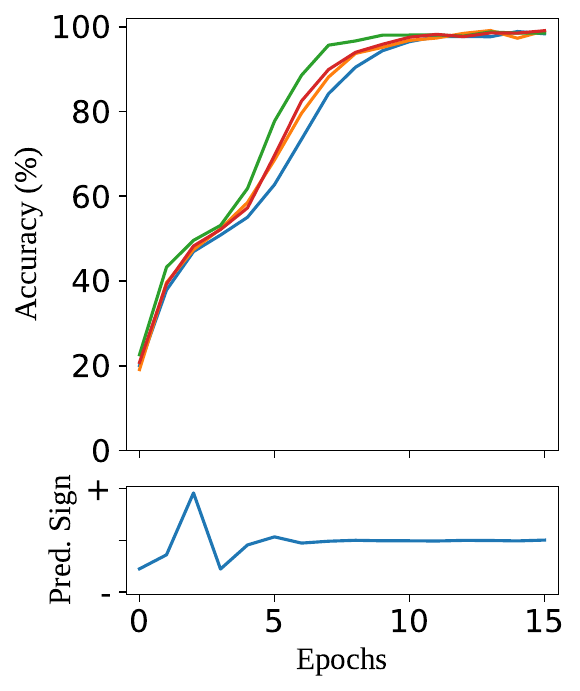} 
    \includegraphics[width=0.51\linewidth]{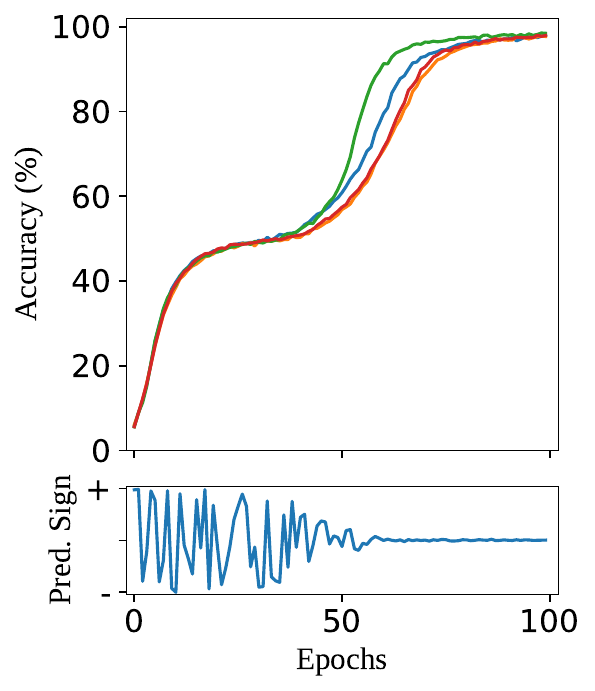}   
\end{tabular}
\caption{Accuracy \emph{vs.} epoch on the nonzero coefficient-from-key task at loop $L=5$ (left) and $L=6$ (right), for four model initializations shown in different colors. The bottom plots show the balance of predicted signs vs. epoch, with $+$ ($-$) indicating $100\%$ ($0\%$) positive signs. Initially the model fluctuates between strongly favoring one sign or the other before more accurately predicting the mix of signs for individual terms.
}
\label{fig:loop56}
\end{figure}

At $L=5$, the best model (out of four initializations, or \textit{seeds}) correctly predicts $43\%$ of the coefficients from the test set after only one epoch. Accuracy is $95\%$ after $7$ epochs, $99\%$ after $16$ epochs and $99.5\%$ after $47$ epochs. 
At $L=6$, the best model (again out of four initializations) correctly predicts $95\%$ of the coefficients in the test set after $66$ epochs, $98\%$ after $88$ epochs and $99.3\%$ after $199$ epochs. We show accuracy as a function of epoch in Figure~\ref{fig:loop56}.

At both $L=5$ and $6$, learning proceeds in two qualitative phases: first the magnitudes of the coefficients are learned, then the signs. At $L=5$, after two epochs, $97\%$ of magnitudes are correctly predicted, but the signs are predicted at near chance level ($50\%$). After $5$ epochs, $99\%$ of magnitudes are predicted correctly, but only $78\%$ of signs are predicted correctly. By epoch $10$, $98\%$ of signs are predicted correctly. For $L=6$, training follows the same pattern, but proceeds more slowly: the model learns the magnitudes of coefficients during the first $20$ epochs. Accuracy then saturates around $50\%$, while the model predicts the magnitudes of coefficients with more than $95\%$ accuracy, but predicts their signs at near chance level. Finally, from epoch $40$ to $70$, the model learns to correctly predict the signs of the coefficients.

Until the sign is learned, the model strongly prefers to predict one sign over the other in each epoch: predictions may for example flip from $98\%$ `$-$' signs in one epoch to $90\%$ `$+$' signs in the next. This preference gradually diminishes with epoch, decreasing to within a few percent of the true proportion of positive and negative signs (which is very close to $50\%$ positive and $50\%$ negative) roughly at the midpoint of the second step (epoch $7$ for $L=5$; epoch $60$ for $L=6$). We indicate this behavior for one representative run in the lower portion of Figure \ref{fig:loop56}. These fluctuations are not restricted to a particular subset of magnitudes; when the true or predicted magnitude is restricted to a given value, similar gradually-decreasing fluctuations are observed.

Furthermore, when models are trained to predict only the magnitudes of nonzero coefficients at $L=6$ (by setting all signs to `+'), the best model can do so at $97.7\%$ accuracy after $50$ epochs. However, when models are trained to predict only the signs of the nonzero coefficients (by setting all magnitudes to `1'), they exhibit random guessing behavior even after 100 epochs. These results suggest that learning the magnitude of the coefficient may be a prerequisite for learning the sign.

In summary, our results indicate that Transformers trained on a small fraction of the symbol can predict coefficients from their keys with very high accuracy.

%%%%%%%%%%%%%%%%%%%%%%%%%%%%%%%%%%%%%%%%%%%%%%%%%%%%%%%%%%%%%%%%%%%%%%%%%%%%%%%
\section{Quad Representation of Symbols}
\label{sec:quads}

At higher loops, a more compact representation of the form factor symbol is necessary for efficient training. For example, the loop $L=7$ symbol has $93$ million nonzero-coefficient elements, almost 19 times as many as the $L=6$ symbol. 
How long might it take to train on such a large symbol?  At loop $L=5$, performing the nonzero coefficient-from-key prediction task to $>99\%$ accuracy takes about 22 passes through the training set and 0.7 hours. At loop $L=6$, the same task takes 11 passes through the training set and over 54 hours. The time to reach $>99\%$ accuracy scales at least linearly with the number of elements. If we assume this scaling continues to $L=7$, it will take at least $54 \times 19 \approx 1000$ hours, or 43 days.

Compressing the data using the quad representation can significantly improve the model training speed by reducing the number of nonzero-coefficient elements to a more manageable $7.3$ million. Under the linear scaling paradigm, we would predict a more manageable 80 hours, or under 4 days.

However, learning coefficients from keys in the quad representation is a harder problem than in the full representation. The quad representation eliminates many of the obvious symmetries in the symbol: it removes both the dihedral symmetry and relations involving up to four final entries. Thus, many potential sources of correlation between the training and test sets are no longer present, and the model is forced to learn more subtle correlations between coefficients and keys in order to correctly predict the coefficients in the test set. Larger models are therefore required for this task. 

Here we train models to predict nonzero coefficients from keys in the quad representation at $L=6$ and $L=7$. For $L=6$, at which there are 391,570 quad keys, we use $4$-layer Transformers with dimension $d=512$ and $8$ attention heads; if we train smaller $2$-layer Transformers with the same dimensions, we are unable to reach above $50\%$ accuracy even after 100 epochs of training. We train on 381,570 key-coefficient pairs and test on the 10,000 held-out elements. 

The training curves for the quad representation exhibit a two-step shape very similar to those for the uncompressed representation: during the first $10$ epochs, only the magnitudes of coefficients are learned and their signs are predicted at chance level. The model achieves an accuracy of $95\%$ in $43$ epochs, which equates to $34$ passes over the compressed training set since an epoch is fixed at $300,000$ examples. The accuracy peaks at $99\%$ after $192$ epochs, equivalent to $152$ passes over the compressed training set. However, the full representation achieves $95\%$ accuracy in $64$ epochs, which equates to slightly less than 4 passes through the uncompressed training set. Thus, although the larger models trained on the quad representation can reach performance benchmarks in fewer epochs than the smaller models trained on the full representations, they in fact take more passes through the training set in order to converge, confirming our intuition that training on the quad representation is a more challenging task.

For $L=7$, even larger models are required. $4$-layer Transformers with dimension $d=1024$ and $16$ attention heads are trained on the entire quad-compressed $L=7$ symbol (i.e., $7.3$ million elements minus the held-out test set of 10,000 elements),
achieving $99.1\%$ accuracy on the test set. Here, training is considerably slower: the signs only begin to be learned after $100$ epochs (\emph{vs.}~$10$ epochs for the $L=6$ symbol). The model reaches $98.5\%$ accuracy in about $400$ epochs, which equates to $120$ million examples, or $16$ passes over the training set. The larger training set partially accounts for this difference in learning speed. 

In order to explore the relationship between  model capacity and training set size, we present in Table~\ref{tab:quad_results} the final overall test-set accuracy for the $L=7$ symbol in the quad representation for different model hyperparameters and training set sizes. We indicate models that fail to achieve at least $90\%$ accuracy in gray. 
Accuracy decreases as training set size decreases, but remains above $94\%$ for all but the smallest model, as long as the models are trained on 3 million examples or more. Such a training set translates to only about $41\%$ of the symbol elements, which suggests that the models still possess significant predictive power even when data is given in the compressed quad representation.

The Transformer’s ability to learn the sign appears to be governed largely by model capacity and training dataset size. Thus, in regions where the model sees enough data and is large enough to consistently learn the sign, accuracy is consistently $>90\%$, while in regions where the model is too small, the sign is not learned, and accuracy stays at or below $50\%$. When we are near a model capacity threshold (for this task, $4$ layers and $d=512$) the model sometimes learns the sign and sometimes does not.

\begin{table}[h]
    \small
    \centering
    \begin{tabular}{lcccccccc}
        \toprule
        \backslashbox{Arch.}{Train. size} & $7.3$M & $7$M & $6$M & $5$M & $4$M & 3M & 2M & 1M \\
        \midrule
        $8$ layers, $d = 1024$ & 98.8\% & 98.7\% &98.2\%  & 97.5\% & 96.7\% & 94.8\% & 90.8\% & \textcolor{gray}{78.2\%} \\
        $8$ layers, $d = 512$ & 96.2\% & 97.4\% & 98.4\% & 96.6\% & 95.3\% & 93.8\% & \textcolor{gray}{88.5\%} & \textcolor{gray}{36.7\%} \\ 
        $6$ layers, $d = 1024$ &98.6\%  & 98.9\% & 98.0\% & 97.9\% & 96.7\% & 94.8\% & 90.3\% & \textcolor{gray}{58.5\%} \\
        $6$ layers, $d = 512$ & 95.2\% & 96.6\% & 96.9\% & 95.8\% & 94.4\% & 94.5\% & \textcolor{gray}{87.9\%} & \textcolor{gray}{34.8\%} \\
        $4$ layers, $d = 1024$ & 99.1\% & 98.9\% & 98.3\% & 97.9\% & 96.6\% & 94.9\% & \textcolor{gray}{89.9\%} & \textcolor{gray}{39.1\%} \\
        $4$ layers, $d = 512$ & \textcolor{gray}{48.5\%} & 96.0\% & 94.1\% & \textcolor{gray}{48.3\%} & 94.6\% & \textcolor{gray}{81.7\%} & \textcolor{gray}{55.3\%} & \textcolor{gray}{33.9\%} \\
       \bottomrule
    \end{tabular}
    \caption{Maximum test-set accuracy after 250 epochs at loop 7 in the quad representation, for various training set sizes as well as numbers of layers and dimensions. The best of two models is shown. All models have 16 heads. (The smallest model occasionally does not emerge from the first plateau, with its accuracy then staying below $50\%$.)}
    \label{tab:quad_results}
\end{table}

%%%%%%%%%%%%%%%%%%%%%%%%%%%%%%%%%%%%%%%%%%%%%%%%%%%%%%%%%%%%%%%%%%%%%%%%%%%%%%%
\section{Model Characterization via Relationship Accuracy}
\label{sec:relations}

The results of the previous experiments strongly suggest that the model leverages certain correlations that are present in the data, such as dihedral symmetry and the final-entry relations described in Section~\ref{ssec:reldef}, in order to more easily extrapolate from the training set into the test set. Additionally, the unusual two-phase accuracy curves that occur in both the full and quad representations warrant further investigation. In this section, we therefore explore how the linear relations behave as a function of epoch in order to better understand how the model learns.

We define an \textit{instance} of a relation as a set of keys and their associated coefficients that obey a given relation. For example, for the relation given in \eqn{integ1_main}, a sample instance at 5 loops is, as in \eqn{integ1_instance}:
\be
C^{c,\underline{a,b},c,a,b,d,c,c,d} + C^{c,\underline{a,c},c,a,b,d,c,c,d} - C^{c,\underline{b,a},c,a,b,d,c,c,d} - C^{c,\underline{c,a},c,a,b,d,c,c,d} = 0
\label{integ1_instance_ALT}
\ee
which corresponds to the following set of keys and coefficients: \{$\tt{cabcabdccd}$: $72$, $\tt{caccabdccd}$: $-88$, $\tt{cbacabdccd}$: $-72$, $\tt{ccacabdccd}$: $56$ \}.

We generate 500 instances of each homogeneous linear relation at loop $L$ and use them to evaluate the performance of a model trained on the coefficient-from-key prediction task as in Section~\ref{sec:coefffromword}. To do so, we randomly generate a set of keys that obey the given relation and then pair them with the corresponding coefficients. We discard and re-generate all such multi-term instances that do not contain at least one nonzero coefficient. In this way, we avoid instances with all zero terms that are trivially satisfied.

The relation instances generated are used only as an auxiliary test set. The training set consists of the full nonzero symbol plus an equal proportion of zeros as in Section~\ref{sec:coefffromword}, while the coefficient-from-key test set is still employed as before. We note that more than 99\% of the nonzero terms in the relation instances appear in the training set. However, we stress that this does not constitute data leakage, as the relation evaluations are auxiliary dataset-level metrics that do not influence training.

The linear relations may relate nonzero terms to the nontrivial zeros, which constitute a very small fraction of all possible zeros. Therefore, when choosing the zeros to be added to the training set, we explicitly select a large proportion of nontrivial zeros. 
The fraction of trivial zeros in the training set is restricted to be $5\%$ of all zeros. 
Prioritizing the nontrivial zeros causes accuracy to decrease slightly on the trivial zeros; however, as trivial zeros are easy to learn and to identify (as per\ Section~\ref{sec:coefffromword}) we can simply manually set the predicted coefficient to zero for any trivial-zero terms in a relation instance.

The model used for this experiment is a 2-layer Transformer with $d=512$ and 8 heads, trained on 9,732,932 elements (i.e., the full $L=6$ symbol plus an equal proportion of zeros), and tested on 100,000 randomly chosen held out elements. After 200 epochs, the model correctly predicts $98.47\%$ of the coefficients in the test set. The learning curves again reveal the familiar two qualitative phases, though training to a given accuracy now takes twice as many epochs (due to the addition of zeros increasing the dataset size by a factor of two) and overall magnitude and sign accuracy both reach $50\%$ accuracy within the first epoch, due to the fact that zeros are learned quickly.

A complete list of the relations we evaluate is given in \ref{app:rels_list}. Here, we only describe the following short relations, which form a representative subset of the different types of relations studied. In the triple and integrability relations, the specified adjacent slots can appear anywhere in the key, while the specified adjacent slots in the final-entry relations must appear at the end of the key (which we denote by $\mathcal{E}$ instead of $F$).

\be
\texttt{triple~0:} \quad F^{a,a,b}+F^{a,b,b}+F^{a,c,b} = 0,
\label{triple}
\ee
\be
\texttt{integ~0:} \quad F^{a,b}+F^{a,c}-F^{b,a}-F^{c,a} = 0,
\label{integ0}
\ee
\be
\texttt{integ~1:} \quad F^{c,a}+F^{c,b}-F^{a,c}-F^{b,c} = 0,
\label{integ1}
\ee
\be
\texttt{final~16:} \quad \mathcal{E}^{b,f}-\mathcal{E}^{b,d} = 0,
\label{final19}
\ee
\be
\texttt{final~17:} \quad \mathcal{E}^{c,d,d}+\mathcal{E}^{c,e,e} = 0,
\label{final20}
\ee
\be
\texttt{final~18:} \quad \mathcal{E}^{d,d,b,d}-\mathcal{E}^{d,b,d,d} = 0.
\label{final21}
\ee

\begin{figure}
\centering
\setlength{\tabcolsep}{-3pt}
\begin{tabular}{ccc}
\includegraphics[width=55mm]{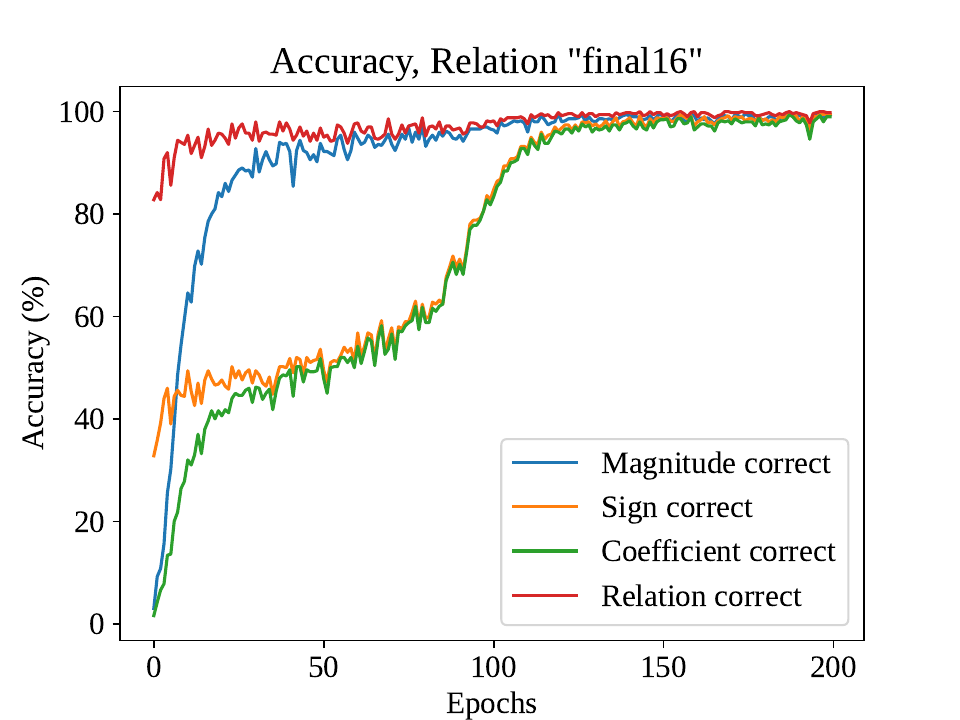} &
\includegraphics[width=55mm]{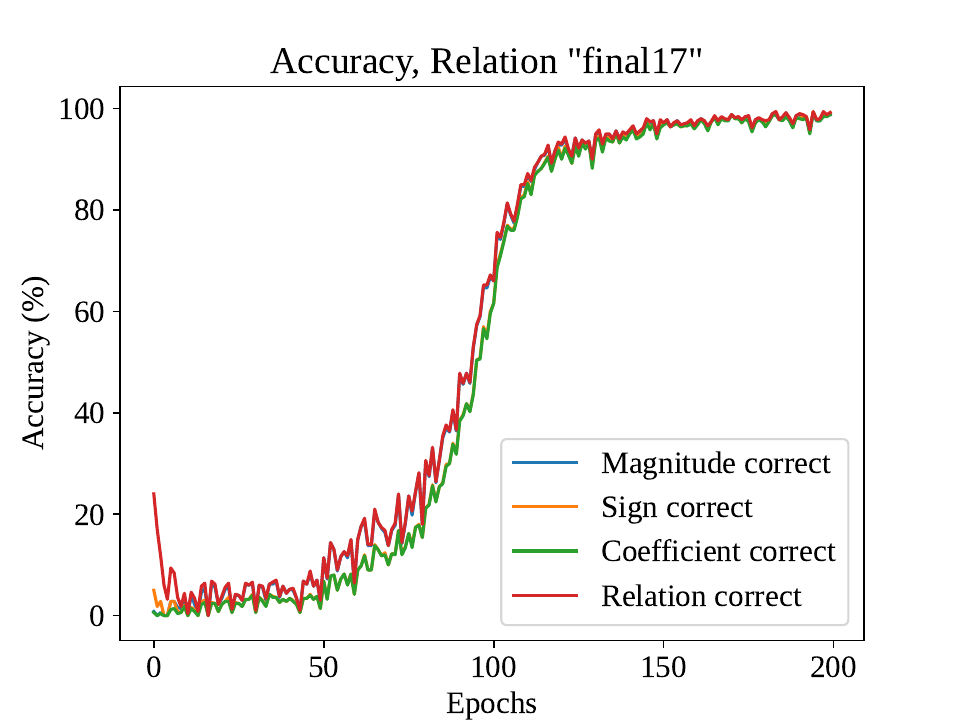} &
\includegraphics[width=55mm]{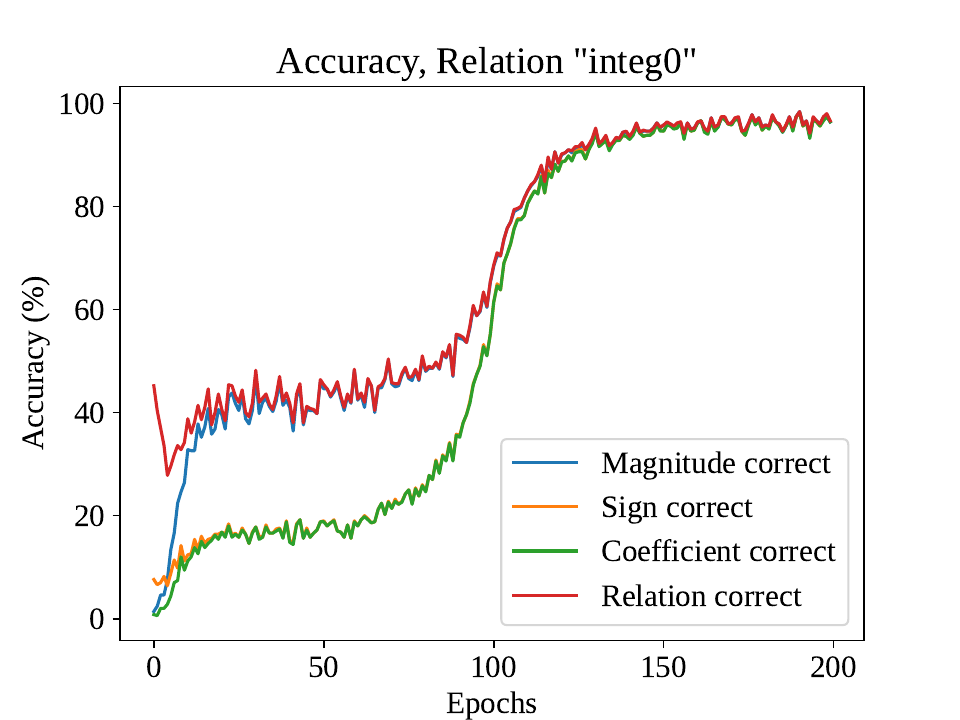} \\  
\includegraphics[width=55mm]{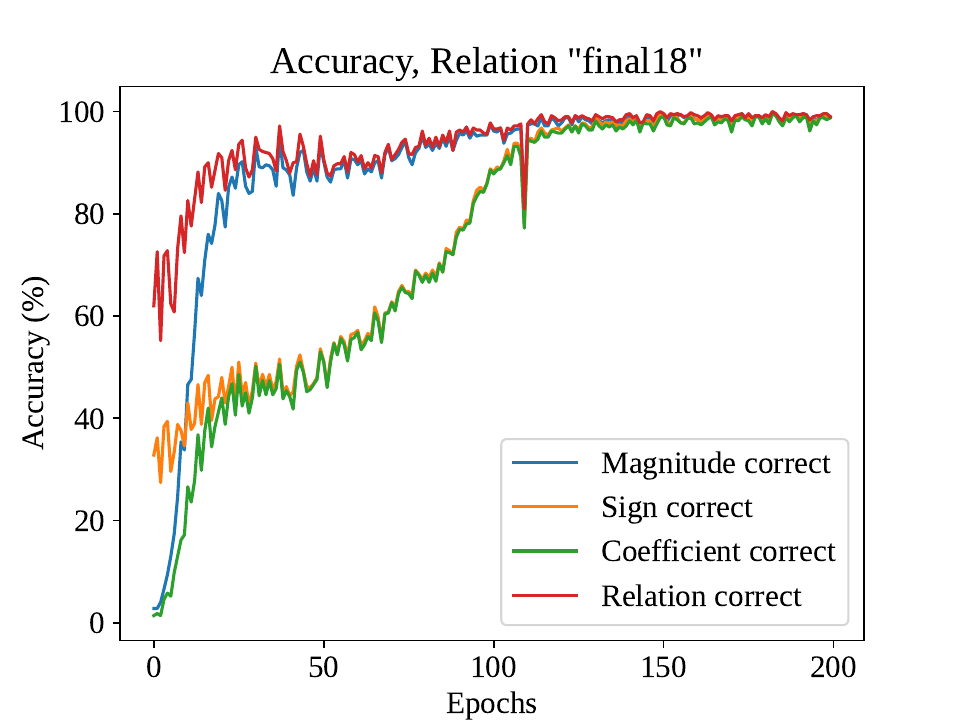} &
\includegraphics[width=55mm]{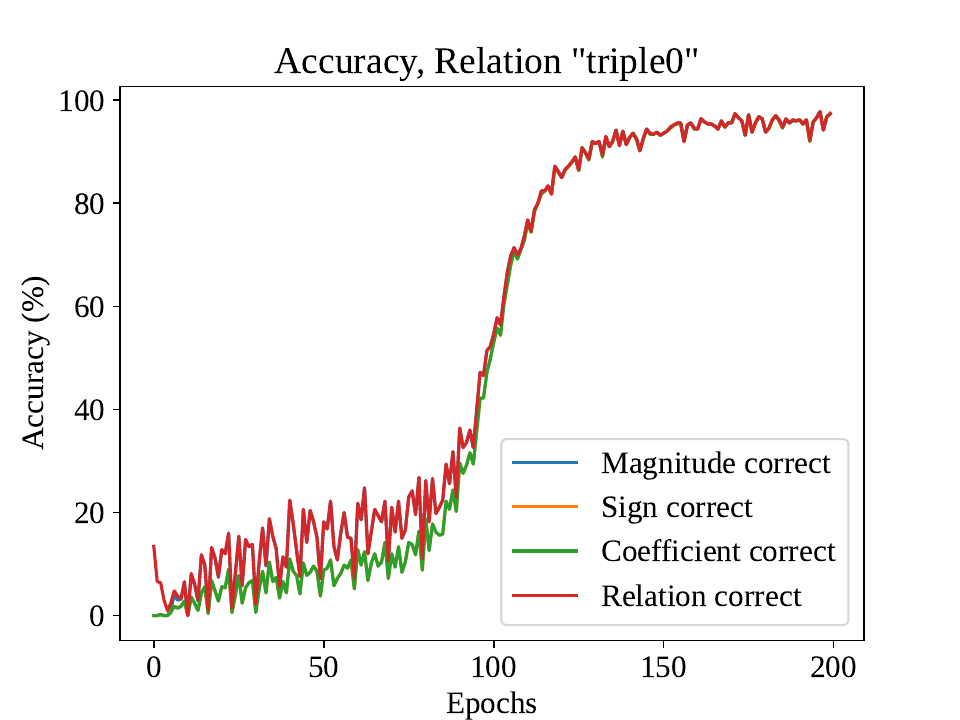} &
\includegraphics[width=55mm]{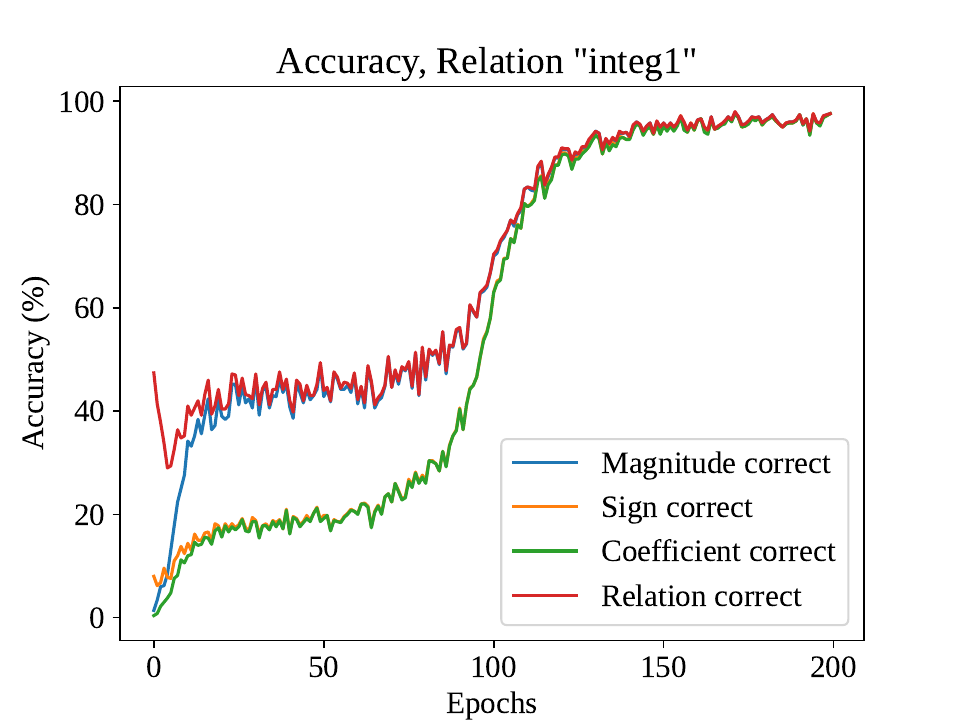} \\
\end{tabular}
\caption{Relation accuracy (red), magnitude accuracy (blue), sign accuracy (yellow), and coefficient accuracy (green) for each of the named relations, grouped by behavior. Relations in Group 1 (left column) are two-term equivalence relations that are consistently satisfied after only a few epochs. Relations in Group 2 (center column) are relations that require at least two coefficients to have different signs, and are not satisfied until the second phase. Relations in Group 3 (right column) are mixed relations, for which some instances decompose into pairs of equivalent terms (as in Group 1 relations) while others do not.}
\label{fig:rel_acc_loop6}
\end{figure}

For each relation, we quote four metrics: 1) whether the coefficients predicted by the model satisfy the given relation, regardless of whether the individual coefficients themselves are correct (red); 2) whether the relation is satisfied and all coefficients in the instance have the correct magnitudes, regardless of their signs (blue); 3) whether the relation is satisfied and all coefficients in the instance have the correct signs, regardless of their magnitudes (yellow); and 4) whether all coefficients in the instance are correct (green). We plot these metrics as a function of epoch in Figure~\ref{fig:rel_acc_loop6}. Similar results for additional relations are provided in \ref{app:rels_list}.

The linear relations can be grouped by structure into three categories that also define their behavior. The six relations discussed in this section contain two examples from each category. 

\begin{itemize}
    \item Group 1 relations, such as \eqn{final19} (\texttt{final 16}) and \eqn{final21} (\texttt{final 18}), are equivalence relations that require two coefficients to have the same magnitudes and signs. For all Group 1 relations, the relation is often satisfied before all the magnitudes are predicted correctly---the model predicts that both coefficients in the instance must be the same before it is able to successfully identify what that coefficient is. While the model learns the magnitudes of both coefficients in the relation instances fairly quickly, the signs of the coefficients are only predicted correctly $50\%$ of the time until the sign is learned; however, both coefficients are consistently predicted to have the same sign. Thus, the sign fluctuations described in Section \ref{sec:coefffromword} (Figure \ref{fig:loop56}) largely respect the Group 1 relations.
    
    \item Group 2 relations, such as \eqn{triple} (\texttt{triple 0}) and \eqn{final20} (\texttt{final 17}), are those in which at least two coefficients must have opposite signs. While the magnitudes of coefficients in these relation instances are learned within a few epochs, accuracy on all relation metrics remains low, but steadily increasing, until the signs are learned. This behavior is also largely dictated by the sign fluctuations shown in Figure \ref{fig:loop56}.

    \item Group 3 relations, such as \eqn{integ0} (\texttt{integ 0}) and \eqn{integ1} (\texttt{integ 1}), are multi-term relations that may be satisfied by two or more pairs of identical coefficients or by a set of related but nonidentical coefficients (e.g., $40\%$ of the generated (\texttt{integ 0}) instances are expressible as pairs of identical coefficients). Under the conditions where the model has a high probability of satisfying the Group 1 relations, the model predictions for the subset of Group 3 relation instances expressible as pairs of identical coefficients will also satisfy the relation. 
\end{itemize}

These properties suggest an explanation for the double plateau behavior: first, the model learns to group elements whose coefficients have the same magnitude; then it learns to correctly predict those magnitudes. The model predictions for the sign fluctuate from epoch to epoch---rather wildly at first---until the second accuracy step is reached. However, these fluctuations consistently respect the Group 1 equivalence relations between elements. During this fluctuation phase, the model also gradually learns the Group 2 relations; i.e., the model learns which coefficients with a given magnitude have the same sign and which do not. These fluctuations get smaller until the sign is eventually learned.

Both the cycle and flip symmetries can be expressed as Group 1 relations relating pairs of terms with identical coefficients. We plot the relation evaluation metrics for these relations in \ref{app:rels_list} and find that they behave similarly to other Group 1 relations. 

We observe a further intriguing manifestation of the dihedral symmetry in the geometry of the embedding layer representation. Specifically, we extract the learned $d$-dimensional embeddings of the input letter tokens from the embedding layer of the Transformer and calculate the angles between them. These embedding vectors obey both the cycle and flip symmetries. We perform such an extraction experiment with a 2-layer Transformer with $d=512$, first at $L=5$ and then at $L=6$.

At $L=5$, the triangles $\triangle{abc}$ and $\triangle{def}$ are approximately equilateral: all angles are within $1.5^\circ$ of $60.0^ \circ$ for $\triangle{abc}$ and within $2.7^ \circ$ of $60.0^ \circ$ for $\triangle{def}$.  This result indicates that the embedding vectors obey the cycle symmetry. Similarly, all angles are within $3.7^\circ$ of $60.0^ \circ$ for triangle $\triangle{abf}$, within $1.5^ \circ$ of $60.0^ \circ$ for $\triangle{bcd}$, and within $3.3^ \circ$ of $60.0^\circ$ for $\triangle{ace}$; the fact that these triangles are approximately similar indicates that the embedding vectors obey the flip symmetry.

At $L=6$, we observe the same phenomenon even more strongly: $\triangle{abc}$ and $\triangle{def}$ are approximately equilateral: all angles are within $1.0^ \circ$ of $60.0^ \circ$ for $\triangle{abc}$ and within $0.6^ \circ$ of $60.0^ \circ$ for $\triangle{def}$, indicating cycle symmetry. Likewise, all angles are within $0.5^ \circ$ of $60.0^ \circ$ for triangle $\triangle{abf}$, within $0.7^ \circ$ of $60.0^ \circ$ for $\triangle{bcd}$, and within $0.8^ \circ$ of $60.0^ \circ$ for $\triangle{ace}$, indicating flip symmetry.

\begin{figure}
\centering
\begin{tabular}{cc}
    \includegraphics[width=0.4\linewidth]{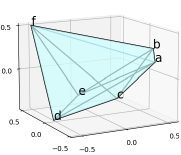}&
    \includegraphics[width=0.4\linewidth]{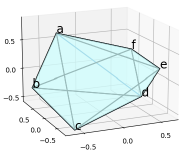}
    
    \\
\end{tabular}
\caption{(Left) The leading three PCA components of token embeddings for a 2-layer Transformer with $d=512$ trained for 50 epochs on $L=5$ data, with zeros included. The leading three PCA components explain $63.56\%$ of variance, and dihedral symmetry is not visible. (Right) The leading three PCA components of token embeddings for a 2-layer Transformer with $d=512$ trained for 200 epochs on $L=6$ data, with zeros included. The leading three PCA components explain $81.76\%$ of variance.  The octahedron exhibits dihedral symmetry.
}
\label{fig:polytope}
\end{figure}

We perform standard linear principal component analysis (PCA) and plot the embeddings of these letter tokens in the space of the three leading PCA components in Figure~\ref{fig:polytope}. Projecting to the leading three components distorts the angles somewhat: at $L=5$, the dihedral symmetry is no longer apparent, while at $L=6$, the cycle and flip symmetries are visually apparent but the octahedron is no longer regular.

%%%%%%%%%%%%%%%%%%%%%%%%%%%%%%%%%%%%%%%%%%%%%%%%%%%%%%%%%%%%%%%%%%%%%%%%%%%%%%%
\section{Mixed-loop Training}
\label{sec:mixed_loop}

To successfully extend the bootstrap program to unseen loops, we must build models that can generalize from lower loops, for which we have the complete symbol, to higher loops, where only a small number of symbol terms may be available.

However, in many AI-for-mathematics applications, Transformers trained exclusively at one input length using absolute position encoding fail to generalize to different input lengths \cite{Nogueira}. Here we face a similar challenge. When using loop $L=6$ data to evaluate a model trained exclusively at $L=5$ (and vice versa) for the task of predicting nonzero coefficients, our baseline models can only attain an accuracy of at most $3\%$ for a variety of model sizes and depths. Many predicted coefficients at unseen loops are nonsensical, such as the string \texttt{`+++'}. Given that our ultimate goal is to predict coefficients of keys at unseen higher loops, this failure of length generalization presents a major limitation that we must overcome.

In many ways, however, this problem goes even beyond simple length generalization. At each subsequent loop, the number of possible values of keys and coefficients both increase appreciably (see Figure \ref{fig:IIueq1Hist}), and it is not clear whether or how the functional form that relates them changes as well. Therefore, while alternative architecture designs, positional encoding schemes, and numerical encoding schemes (which we explore in ~\ref{app:ablations}) may be helpful for this task, a much more comprehensive strategy will likely be needed.

\begin{figure}[bp]
\centering
\begin{tabular}{cc}
    \includegraphics[width=0.4\linewidth]{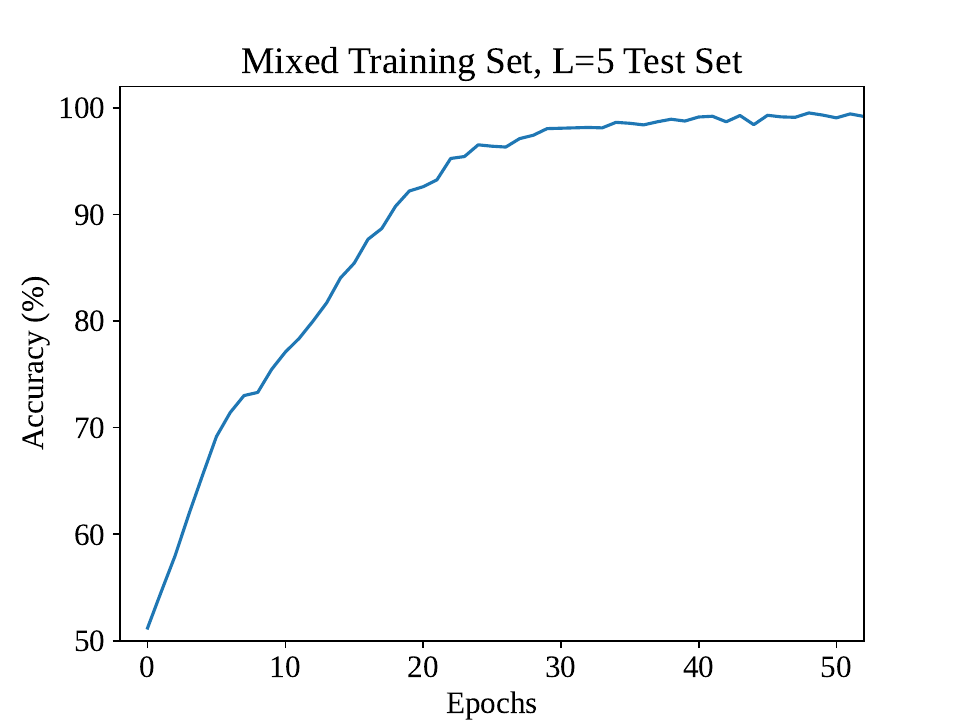} & 
    \includegraphics[width=0.4\linewidth]{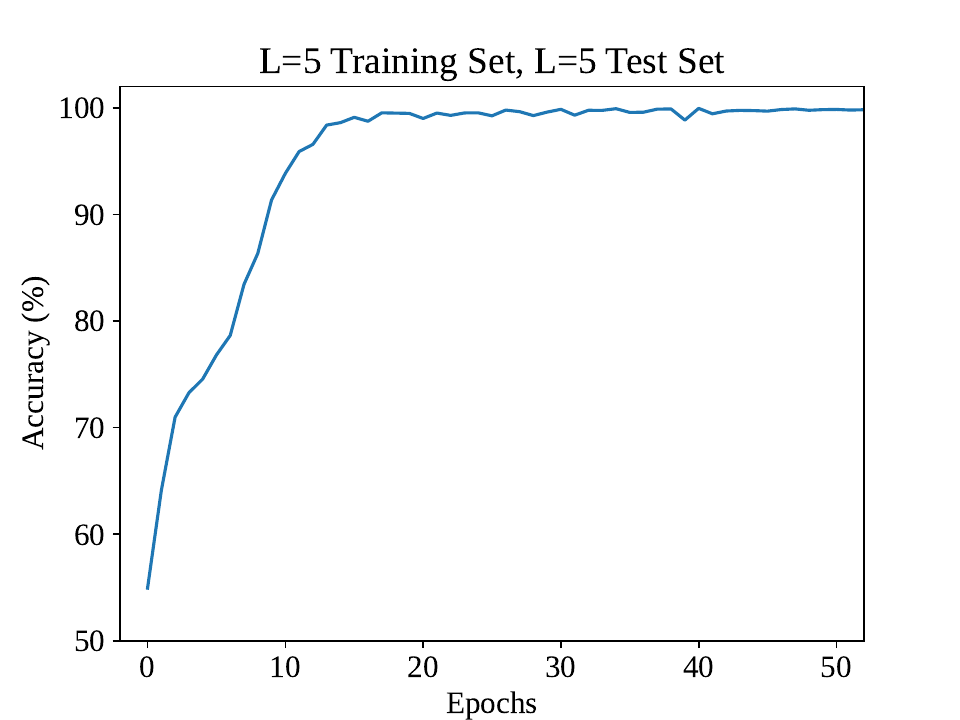} \\
    \includegraphics[width=0.4\linewidth]{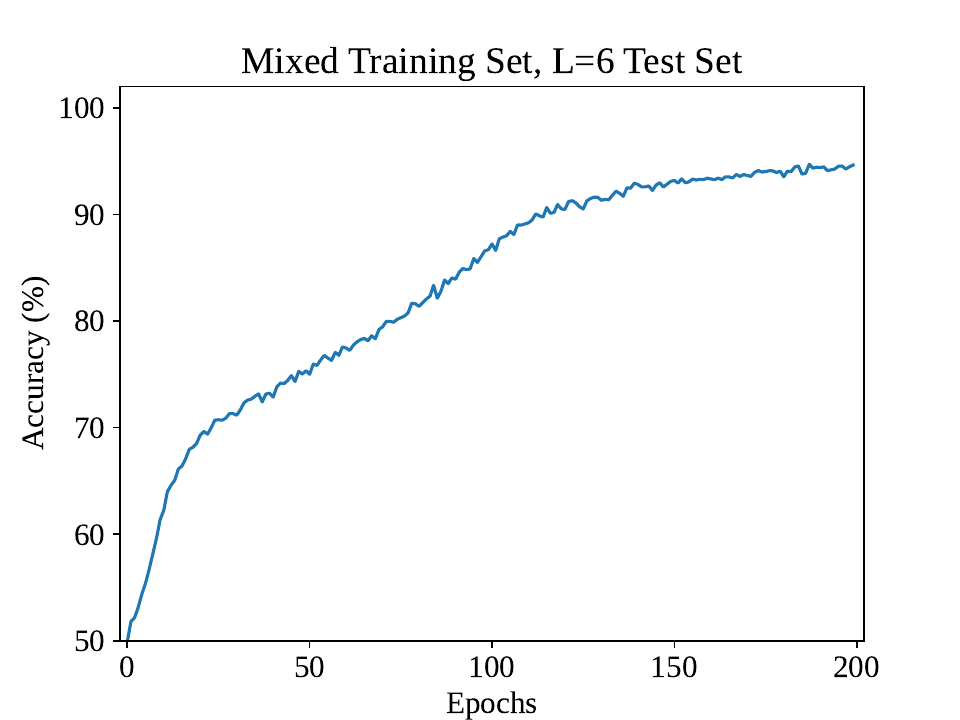} &
    \includegraphics[width=0.4\linewidth]{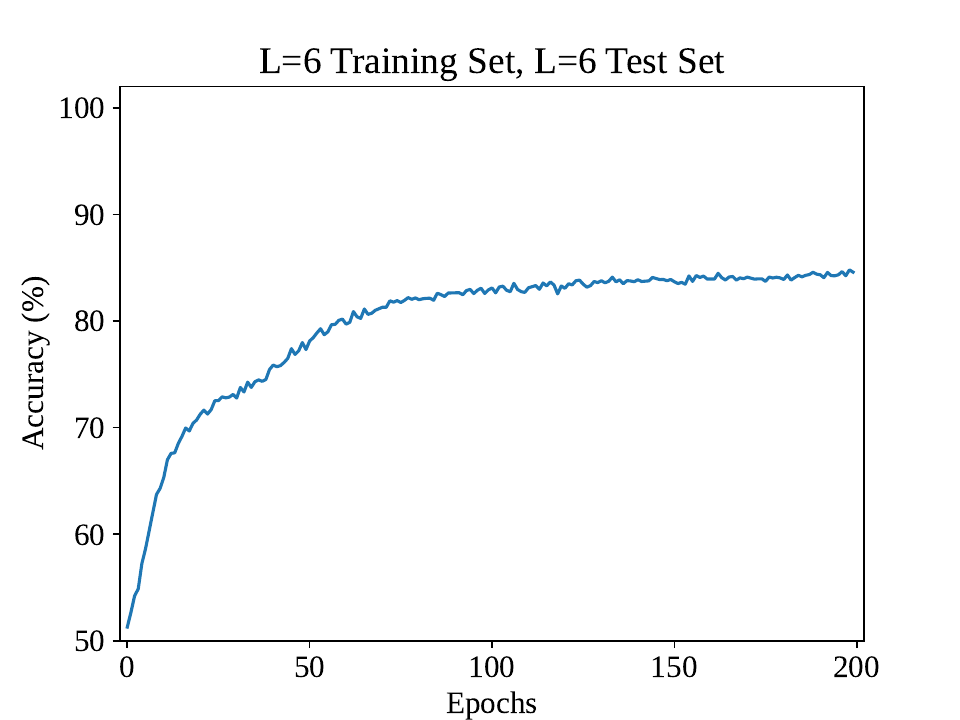}
\end{tabular}
\caption{Accuracy \emph{vs.} epoch for models trained on the mixed and per-loop  training sets, evaluated on the per-loop test sets. In all cases, accuracy starts at 50\% because zeros are learned in the first epoch. Models trained on the mixed training set take almost exactly half the number of epochs to reach performance benchmarks for $L=5$ as models trained on the dedicated $L=5$ training set, corresponding to approximately equal performance of these two models. However, models trained on a small subset of the $L=6$ symbol mixed with the full $L=5$ symbol are able to generalize to the $L=6$ test set much better than models trained on the $L=6$ symbol alone.}
\label{fig:mixed_loops}
\end{figure}

As a first attempt to address this issue, we train Transformers to predict coefficients using an even proportion of $L=5$ and $L=6$ data. The full nonzero $L=5$ symbol (263,880 elements) is first combined with an equal proportion of zero-coefficient elements at $L=5$. Another 263,880 nonzero elements are then drawn from the $L=6$ symbol (representing $5\%$ of the $L=6$ symbol) and augmented with a roughly equal amount of zero-coefficient elements at $L=6$. The zero-sampling is done na\"ively, as in Section~\ref{sec:coefffromword}, rather than in the nontrivial-zero-biased manner of Section~\ref{sec:relations}. Training and test sets are constructed for $L=5$ and $L=6$ separately, and they are then merged to create \textit{mixed-loop} training and test sets. Each loop-specific training set for both $L=5$ and $L=6$ contains 517,760 examples, while each loop-specific test set contains 10,000 examples; every set is an equal mix of zero- and nonzero-elements. Additionally, we create a larger $L=6$ training set that is the same size as the mixed-loop set (1,035,520 examples, roughly 10\% of the $L=6$ symbol), in order to evaluate whether the effects of mixed-loop training can be explained by the difference in training set size. 

We train a model with $2$ layers, $d=512$, and $8$ attention heads in both the encoder and decoder for 200 epochs, evaluating on both the mixed-loop test set and the individual $L=5$ and $L=6$ test sets. The model again exhibits two-phase learning behavior in all cases, as shown in Figure~\ref{fig:mixed_loops} (which displays results for all but the larger $L=6$ training set). We measure the performance by reporting three epochs,
\begin{enumerate}
    \item the epoch at which the test-set magnitude accuracy first exceeds $90\%$,
    \item the midpoint epoch of the plateau step, which is when the model's overall accuracy first reaches the average between its final-state accuracy and $75\%$,
    \item the epoch at which the overall test-set accuracy first exceeds $90\%$,
\end{enumerate}   
as well as the best overall test-set accuracy after 200 epochs.

\begin{table}[bp]
    \small
    \centering
    \begin{tabular}{llccccc}
        \toprule
        Train & Eval & \CellWithForcedBreak{First Mag. Acc. \\ > 90\% [Epoch]} & \CellWithForcedBreak{Midpoint of \\ Step [Epoch]} & \CellWithForcedBreak{First Total Acc. \\ > 90\% [Epoch]} & \CellWithForcedBreak{Best Acc.\\Epoch 200} \\
        \midrule
        
        Mix& $L=5$ & 6 & 16 & 18 & 99.82\% \\
        $L=5$&  $L=5$ & 3 & 9 & 10 & 99.9\% \\
        Mix & $L=6$ &  25 & 88 & 112 & 94.6\% \\ 
        $L=6$& $L=6$ & 20 & 51 & N/A & 84.5\% \\
        $L=6$, Large& $L=6$ & 17 & 92 & 100 & 97.67\% \\
       \bottomrule
    \end{tabular}
    \caption{Model learning dynamics for mixed training: the epoch at which the test-set magnitude accuracy first exceeds 90\%, the midpoint of the plateau step, the epoch at which the overall test-set accuracy first exceeds 90\%, and the best overall test-set accuracy after 200 epochs. All training hyperparameters (architecture, initialization seed, etc.) are kept the same; only the training and test sets are changed between runs.
    }
    \label{tab:mix_scan}
\end{table}

In Table \ref{tab:mix_scan}, we compare the model performance when trained on each loop-specific training set to its performance when trained on the mixed training set. The dedicated $L=5$ model reaches all performance benchmarks for each loop in almost exactly half the number of epochs as the mixed-loop model. Accounting for the fact that the mixed-loop training set is exactly twice the size of the individual $L=5$ and $L=6$ training sets, this result suggests that the mixed-loop model solves the prediction problem for $L=5$ in almost exactly the same amount of time as it would without the addition of $L=6$ data. On the other hand, the mixed-loop model learns the $L=6$ magnitudes in approximately the same number of epochs as the dedicated $L=6$ model, corresponding to reaching this benchmark (i.e., First~Mag.~Acc.~> 90\%) in half as many iterations through the $L=6$ training set. The mixed-loop model also exceeds $\sim 90\%$ overall accuracy, whereas the dedicated $L=6$ model does not. However, much of this effect may be explained by the dataset size: when a model is trained on the larger $L=6$ dataset that is the same size as the mixed set, the results appear highly similar to those we see when using the mixed training set.

These results are nonetheless rather encouraging: while adding a subset of $L=6$ data to $L=5$ does not lead to improvement on $L=5$ tasks, adding the full $L=5$ symbol to a subset of the $L=6$ symbol does appear to improve performance on $L=6$ tasks. This suggests that features learned at lower loops can be employed to enhance the predictive power of a small number of symbol elements at higher loops (for which we may only be able to determine a handful of coefficients \emph{a priori}). 

%%%%%%%%%%%%%%%%%%%%%%%%%%%%%%%%%%%%%%%%%%%%%%%%%%%%%%%%%%%%%%%%%%%%%%%%%%%%%%%
\section{Steps Toward Predicting the Next Loop}
\label{sec:strikeout}

Although mixed-loop training may allow us to better generalize from only a small number of symbol elements at unseen loops, we still wish to find ways to predict these elements at loops for which we do not have coefficient information. 

As a first attempt at solving this task, we now consider a different problem setup. Instead of predicting unknown symbol elements at a given loop order, we would like to obtain the loop $L$ symbol from the coefficients at loop $(L-1)$. In other words, for any element at loop $L$, we want to recover its coefficient from the coefficients of a list of \emph{parent} elements at one loop lower that are related by having similar strings of letters in their keys.

The keys of elements at loop $L$ are sequences of $2L$ letters, whereas the keys at loop $L-1$ are only $(2L-2)$ letters long. We therefore define the \emph{strike-two parents} of a given key at loop $L$ as the keys from loop $(L-1)$ which are created by simply striking out two letters from the key. For instance, the six strike-two parents of the loop $L=2$ key \texttt{aacf} are 
\be
\texttt{\cancel a\cancel acf=cf}, \quad \texttt{\cancel aa\cancel cf=af}, \quad \texttt{\cancel aac\cancel f=ac}, \quad \texttt{a\cancel a\cancel cf=af}, \quad \texttt{a\cancel ac\cancel f=ac}, \quad \texttt{aa\cancel c\cancel f=aa},
\label{strike-two-example}
\ee
which are keys at $L=1$. In general, there are $\binom{2L}{2} = L(2L-1)$ strike-two parents of any key at loop $L$.

The dataset for this experiment is constructed by first selecting certain elements at loop $L$. For each key, we then construct the list of strike-two parents at loop $L-1$ and list their coefficients in the \textit{strikeout order}, i.e., the coefficient corresponding to striking the first two letters from the key is ordered first in the coefficient list, the coefficient corresponding to striking the first and third letters from the key is ordered second in the list, etc. For the above example at $L=2$, the coefficient to be predicted is $C^{aacf} = 0$, and the ordered list of its strike-two parent coefficients is \texttt{[0, -2, 0, -2, 0, 0]}. Many of the parent coefficients are zero due to the nature of the strike-out. For example, pairs of letters that are not allowed to be adjacent can become adjacent in the parent keys after a letter in between is removed.

In these experiments, $4$-layer Transformers with $d=512$ and 8 heads are trained to predict nonzero coefficients at $L=6$ from the coefficients of their $L=5$ parents. Model inputs are sequences of $\binom{12}{2} = 66$ parent coefficients at $L=5$ given in strikeout order, while the targets are single coefficients at $L=6$.

From the $L=5$ and $L=6$ symbols, we can create $4.9$ million examples ($L=5$ parents and $L=6$ coefficients), but this dataset includes many duplicates. In fact, each example is duplicated $6.4$ times on average, primarily due to the dihedral symmetry. To avoid contamination between the training and test set, we restrict the dataset to the 767,500 unique examples, split into 757,500 training and 10,000 test examples.

We report our results in the first row of Table~\ref{tab:base_ablation}. After $500$ epochs, the model predicts $98.1\%$ of the test examples. Learning is fast: $90\%$ accuracy is achieved after $20$ epochs, and $95\%$ after $80$ epochs. In these experiments, the magnitudes and signs of coefficients are learned simultaneously. In other words, no two-phase learning dynamics are observed, in contrast to previous experiments.

Our results suggest that there may exist learnable formulas for computing coefficients at $L=6$ from their strike-two parents at $L=5$. We are not, so far, capable of explicitly recovering these formulas, but additional ablation experiments may shed light on some of their features. We perform some such experiments in the remainder of this section, and give further results in~\ref{app:recurrence}.

First, we investigate whether we can predict the $L=6$ coefficient from a smaller set of $L=5$ parents. One way is to only strike letters that are no more than $k$ positions away from each other in the $L=6$ keys. For example, with $k=1$ we only strike adjacent letters; for $k=2$ we only strike letters that are either adjacent or separated by one additional letter, etc. 
The number of parents remaining in this reduced set is $2kL-\frac{k(k+1)}2$. At $L=6$, the $66$ parents available at the maximum $k=12$ are reduced to only $11$ parents for $k=1$. We construct the reduced dataset for $L=6$ with $k=1,2,3,5$, remove duplicates, and present the parent coefficients to the model in the strikeout order.

Our models predict $98.3\%$, $98.4\%$, $98.1\%$ and $94.3\%$ of test examples for $k=5$, $3$, $2$ and $1$, respectively. Predicting from $21$ parents ($k=2$), instead of the full $66$, has little impact on model performance, suggesting that the majority of coefficients at higher loops may be learnable using only a limited set of parent coefficients at lower loops.

\begin{table}[h]
    \small
    \centering
    \begin{tabular}{lccc}
        \toprule
          &   Accuracy & Magnitude accuracy & Sign accuracy \\
        \midrule
        Strike-two, all parents  & 98.1\% & 98.4\% & 99.6\% \\
        Strike-two, $k=5$ & 98.3\% & 98.6\% & 99.7\% \\
        Strike-two, $k=3$ & 98.4\% & 98.7\% & 99.7\% \\
        Strike-two, $k=2$ & 98.1\% & 98.3\% & 99.5\% \\
        Strike-two, $k=1$ & 94.3\% & 95.2\% & 98.5\% \\
        \midrule
        Randomly shuffled parents, all parents  & 95.2\% & 99.1\% & 96.3\% \\
        Randomly shuffled parents, $k=2$ & 93.5\% & 98.1\% & 95.0\%  \\
      
        Sorted parents, $k=5$  & 93.9\% & 95.4\% & 97.9\% \\
        \midrule
        Parent magnitudes only, all parents & 81.8\% & 98.4\% & 83.2\% \\
        Parent signs only, all parents  &93.3\% & 93.5\% & 99.0\% \\
        Parent signs only, all parents, sorted  & 0.8\% & 61.0\% & 1.6\% \\
        \bottomrule
    \end{tabular}
    \small
    \vspace{.2em}
    \caption{Overall, magnitude, and sign accuracy for the cross-loop strike-out experiments described in the text. Best of four seeds, trained for about $500$ epochs.}
    \label{tab:base_ablation}
    \end{table}

Next we experiment with the order of the strikeout parents. Surprisingly, even if parent coefficients (i.e., model inputs) are randomly shuffled, the model can still achieve $95.2\%$ accuracy ($93.5\%$ with the $21$ parents for $k=2$). When parents are sorted by increasing order in their numerical values, the model achieves $93.9\%$ for $k=5$. This result suggests that the ability of Transformers to compute coefficients at $L=6$ from $L=5$ is mostly unaffected by permutations of the $L=5$ coefficients.
However, the result may be an artifact of the way the strikeout experiment is constructed. Because certain letter adjacency conditions lead to zero coefficients, an ordered list of parents implicitly encodes some information about the original letter structure of the key. From previous experiments, we know that information about the letter structure of the key can be used to reconstruct the coefficient. In future strikeout experiments, we may wish to take this effect into account in order to better model the relationships between elements across loop orders.

Finally, we modify the values of the parent coefficients themselves in two different ways: (1) we set the signs of all parents to `+' before removing duplicates, thereby retaining only the magnitude information; or (2) we provide only the signs of the parent coefficients as inputs to the model, i.e., each parent coefficient is encoded as $-1$, $+0$ or $+1$.
In both cases, all modified parent coefficients are still presented in the strikeout order.

In case (1), models trained on only the magnitudes of the parents are able to recover the magnitudes of the target coefficients with $98.4\%$ accuracy, about the same level as models trained on unmodified parent coefficients. However, the sign of the target coefficient proves harder to learn when the model cannot see the signs of the parents (dropping from $99.6\%$ to $83.2\%$ accuracy).

In case (2), the models can achieve an overall accuracy of $93.3\%$, and correctly predict the sign in $99\%$ of the test cases. However, if we additionally shuffle or sort these parents in ascending order (i.e., all $-1$s, all 0s, then all $+1$s), we find that the model is totally unable to learn. In other words, we can drastically reduce information about either the values of the strikeout parents or their ordering and still recover the full coefficient, but we cannot do both simultaneously.

The fact that coefficients can be reconstructed reliably from their strike-two parents, despite the fact that symmetries such as dihedral symmetry are removed, suggests that a closed-form solution for computing coefficients from their parents may exist. It is likely that some amount of redundant information exists in the set of strikeout parents, as we are able to reconstruct the coefficient in a number of scenarios when the set of parents is severely altered. We plan to investigate such cross-loop relationships further in future work.

%%%%%%%%%%%%%%%%%%%%%%%%%%%%%%%%%%%%%%%%%%%%%%%%%%%%%%%%%%%%%%%%%%%%%%%%%%%%%%%
\section{Conclusions and Future Work} \label{sec:conclusions}

In this study, we have shown that a Transformer model is able to successfully predict the coefficients of elements in the symbols for scattering amplitudes in $\cN=4$ planar super Yang-Mills theory. Below we summarize the key findings of our work. 

Our models learn in a two-phase fashion, first achieving very high accuracy on the coefficient magnitudes and then learning their signs. The distribution of predicted signs exhibits large fluctuations that gradually decrease as the sign is learned.  The models cannot learn the signs of coefficients without the magnitude information, while they can learn magnitudes without sign information.

Transformers perform very well even when the data is compressed into the quad representation, where many trivial correlations between terms are removed. Enough information remains for a (larger) Transformer to successfully reconstruct coefficients from keys. This result bodes well for our ability to move to higher loops, as the space of possible coefficients and keys becomes quite large beyond $L=6$.

To study the learning dynamics, we have evaluated the models' performance on the linear relations. The relations can be classified into three groups based on whether they require coefficients to be identical, to have opposite signs, or some combination of the two. We propose a likely explanation for the two-phase behavior by assessing which of the known linear relations are satisfied at each epoch: the model first learns to group many terms with the same coefficient magnitude; then it learns what those magnitudes are; next it learns which coefficients with a given magnitude have the same sign and which have a different sign; finally it learns the true signs. Additionally, one of the simplest symmetries of the symbol, its dihedral symmetry, can be seen geometrically in the embedding layers of the Transformer.

We have also trained models on mixed-loop data. Augmenting a small percentage of training examples at one loop with a substantial fraction of the symbol at lower loops leads to faster convergence and higher accuracy on a test set at the higher loop. This performance is particularly encouraging, as it suggests that only a relatively small number of coefficients may need to be provided at unseen loops in order for Transformers to successfully predict the rest of the symbol.

In our second set of experiments, namely the cross-loop strike-out approach, we have shown that a model can predict the coefficients at loop $L$ using only a small subset of related coefficients at loop $(L-1)$. Coefficient information from the lower loop can be scrambled or severely degraded (but not both at once), without hindering the ability to reconstruct the target coefficient.

Transformers are well-studied, but they are not the only architecture with the potential to learn the structure of scattering amplitudes. In \ref{app:archs}, we train a Long-Short-Term Memory recurrent neural network (LSTM)~\cite{HochSchm97} and a Gated Recurrent Unit (GRU)~\cite{ChoGRU} on the $L=5$ symbol. In both cases, we are able to attain comparable performance to a Transformer. It may be interesting to explore other types of architectures as well.

In the future, we plan to train on more complicated objectives, with the goal of developing a model that encodes many types of information from multiple loops. This evokes the concept of a ``foundation model'': by training a large, multitask model with information about a number of relevant concepts, we hope to better characterize the complex recurrences and relations present in the symbol in order to generalize to unseen loops. This is a challenging domain-generalization task, as the distribution of possible keys and coefficients changes substantially between loop orders.

The methods in this paper can be applied straightforwardly to other SYM problems where there is multi-loop data encoded by symbols. Additional challenges arise when considering a similar approach to amplitudes in QCD. While many of the generalized polylogarithms appearing in SYM also appear in QCD, the QCD result does not have uniform weight $2L$, but generally has all weights from $2L$ down to zero (rational functions). The lower-weight terms typically have complicated prefactors that are rational functions of the kinematic variables. Also, the behavior in physical limits is not understood as well in QCD as in SYM. A better understanding of how to organize and bootstrap QCD amplitudes would clearly benefit any machine-learning approaches to determining them.

Ultimately, our goal is to build machine-learning models capable of computing amplitudes analytically {\it ab initio}. In addition to the potential spin-offs for collider physics, understanding even simplified scattering amplitudes to all loop orders would give a remarkable new window into quantum field theory.

%%%%%%%%%%%%%%%%%%%%%%%%%%%%%%%%%%%%%%%%%%%%%%%%%%%%%%%%%%%%%%%%%%%%%%%%%%%%%%%

%%%%%%%%%%%%%%%%%%%%%%%%%%%%%%%%%%%%%%%%%%%%%%%%%%%%%%%%%%%%%%%%%%%%%%%%%%%%%%%
\begin{ack}

The data that support the findings of this study are openly available at the following URL/DOI: \href{https://10.5281/zenodo.11218272}{https://10.5281/zenodo.11218272} \cite{dixon_2024_11218272}. A prior version of this work was presented in a non-archival form at the 2023 NeurIPS Machine Learning for the Physical Sciences workshop \cite{ML4PS}. We thank Yang-Hui He, Romuald Janik, Matt Schwartz, Jesse Thaler, Dmitris Papailiopoulos, Jordan Ellenberg, Gary Shiu, and Yiqiao Zhong for fruitful discussions. We would like to thank Phil Wang (``lucidrains'') for use of his implementation of the RoPE and xPos relative position encodings. The work was supported in part by the U.S. Department of Energy (DOE) under Award No.~DE-FOA-0002705, KA/OR55/22 (AIHEP). LD and TC are additionally supported by the U.S. Department of Energy Award No.~DE-AC02-76SF00515. MW was supported by the research grant 00025445 from Villum Fonden.

\end{ack}

%%%%%%%%%%%%%%%%%%%%%%%%%%%%%%%%%%%%%%%%%%%%%%%%%%%%%%%%%%%%%%%%%%%%%%%%%%%%%%%

\section*{References}

\bibliography{main}
\bibliographystyle{JHEP}

%%%%%%%%%%%%%%%%%%%%%%%%%%%%%%%%%%%%%%%%%%%%%%%%%%%%%%%%%%%%%%%%%%%%%%%%%%%%%%%
\appendix
\newpage

\section{Mapping Generalized Polylogarithms to Symbols} \label{app:symb_map}

One of the major difficulties in computing multi-loop amplitudes is the complexity of the multivariate transcendental functions that are encountered. At one loop, besides the logarithm, only one special function appears, namely the classical dilogarithm
\be
   {\rm Li}_2(x) = - \int_0^x \frac{dt}{t} \ln(1-t)
   = \int_0^1 \frac{dt}{t} \int_0^{t} \frac{dt'}{1-t'} \,.
\ee
We define the {\it weight} of a polylogarithmic function as the number of non-trivial integrations over rational functions it requires. Thus, the classical dilogarithm has weight two, because there are two such integrations. At $L$ loops, functions with weight up to $2L$ are encountered, including generalized polylogarithms~\cite{Goncharov:1998kja,Goncharov:2001iea}, defined recursively by
\be
G(a_1,a_2,\ldots,a_n,x) = \int_0^x \frac{dt}{t-a_1} G(a_2,\ldots,a_n,t),
\qquad G(\vec{0}_n,x) = \frac{1}{n!} \ln^n x,
\label{genpoly}
\ee
with $G=1$ for $n=0$. Such functions are quite well understood mathematically, and that understanding has fueled much of the recent progress in multi-loop scattering amplitudes. 

In more complex processes, generalized polylogarithms are not enough, and elliptic polylogarithms~\cite{Brown:2011wfj} and functions beyond elliptic are needed;  see e.g.~ref.~\cite{Bourjaily:2022bwx} and references therein. Fortunately, the three-gluon form factor in SYM can be expressed in terms of the generalized polylogarithms~\eqref{genpoly}, with a weight that is precisely $2L$ at loop order $L$.

Consider a polylogarithmic function $P$ with weight $n$. It is simplest to describe it iteratively via its derivatives, or total differential, which can be expressed in terms of a number of weight $n-1$ functions $P^{s_k}$:
\be
dP = \sum_{s_k\in\cL} P^{s_k} \, d \ln s_k \,,
\label{dP}
\ee
where $\cL$ is the {\it symbol alphabet} containing {\it letters} $s_k$, which are functions of the underlying kinematical variables. 

One can iterate \eqn{dP} $n$ times, defining the weight $n-2$ functions $P^{s_j,s_k}$ by
\be
dP^{s_k} = \sum_{s_j\in\cL} P^{s_j,s_k} \, d \ln s_j \,,
\label{dPs}
\ee
and so on, finally reaching the {\it symbol}~\cite{Goncharov:2010jf} of $P$, $\cS[P]$, denoted by
\be
\cS[P] \equiv \sum_{s_{i_1},\ldots,s_{i_n}\in\cL} P^{s_{i_1},\ldots, s_{i_n}} \,
s_{i_1} \otimes \cdots \otimes s_{i_n} \,.
\label{symbP}
\ee
Here $P^{s_{i_1},\ldots, s_{i_n}}$ is an $n$-fold tensor of weight-$0$ polylogarithms, i.e., rational numbers, which we denote by $C^{s_{i_1},\ldots, s_{i_n}}$ in the main text. The $\ln$'s associated with the $d\ln s_k$'s are by convention omitted from the tensor product defining the symbol.

For the three-gluon form factor considered in this article, the symbol alphabet is 
\be
\cL_{\tt 3gFF} = \{a,b,c,d,e,f\} \,,
\label{F3alphabet_appendix}
\ee
where the six letters read
\be
a = \sqrt{\frac{u}{vw}}, \quad b = \sqrt{\frac{v}{wu}}, \quad c = \sqrt{\frac{w}{uv}}, \quad
d = \frac{1-u}{u}, \quad e = \frac{1-v}{v}, \quad f = \frac{1-w}{w},
\label{F3letters}
\ee
and the $(u, v, w)$ are rational functions of the kinematical variables: 
\be
u = \frac{s_{12}}{q^2}, \quad v = \frac{s_{23}}{q^2}, \quad w = \frac{s_{31}}{q^2}
\label{F3uvw}
\ee
with $s_{ij} = (p_i + p_j)^2$. The $p_i$'s are the relativistic four-momenta associated with the three on-shell external states, i.e., the three gluons, and $q^2 = s_{123} = s_{12}+s_{23}+s_{31}$.

Given the arbitrary nature of ordering the $p_i$'s, the form factor $\cF_{\tt 3gFF}$ is invariant under a dihedral symmetry group $D_3$ (which is isomorphic to the full permutation group $S_3$).  This group is generated by two transformations which act on the symbol letters as
\be
\text{cycle: } \{a, b, c, d, e, f\} \to \{b, c, a, e, f, d\} \
\text{, and flip: } \{a, b, c, d, e, f\} \to \{b, a, c, e, d, f\}. 
\ee 

Since the symbol involves taking $n$ derivatives, it does not capture all the information in the function. Certain boundary conditions are also required when integrating back up. However, the symbol contains almost all of the critical information, thus serving as a scaffold for determining the full function.

%%%%%%%%%%%%%%%%%%%%%%%%%%%%%%%%%%%%%%%%%%%%%%%%%%%%%%%%%%%%%%%%%%%%%%%%%%%%%%%
\section{Model Evaluation on the Full List of Linear Relations} \label{app:rels_list}

In this appendix, we provide a list of linear relations that the form factor symbol satisfies. 
We also provide results for the accuracy with which the relations hold as a function of training epoch, similar to the discussion in Section \ref{sec:relations} but for 
additional relations.  

We recall from Section \ref{ssec:symbols} that there are {\bf adjacency} and {\bf prefix/suffix} rules that strongly limit which keys can have nonzero coefficients.  Beyond those conditions, there is a large set of linear relationships. We now describe the ones we evaluate during model training.  These linear relationships can be grouped into three classes: the triple-adjacency relation, the integrability relations, and the multiple-final-entries relations. We list the relations we consider without providing justification; see~\cite{Dixon:2020bbt,Dixon:2022rse} for details about the underlying physics.

\begin{itemize}
    \item \textbf{Triple Adjacency:}
    There is \textbf{one} triple adjacency relation (up to dihedral transformations). It relates three terms at a time, and the specified adjacent slots can appear anywhere in the key. When evaluated on the symbol at a certain loop, $F$ can be identified with any of the polylogarithms $P$ described in \ref{app:symb_map}.  
        \be
        \texttt{triple 0:} \quad F^{a,a,b}+F^{a,b,b}+F^{a,c,b} = 0.
        \label{triple_}
        \ee

    \item \textbf{Integrability:}
    There are \textbf{three} different integrability relations, with the longest one relating 14 terms at a time. Again, the specified adjacent slots can sit anywhere in the key. 
        \be
        \texttt{integ~0:} \quad F^{a,b}+F^{a,c}-F^{b,a}-F^{c,a} = 0,
        \label{integ0_}
        \ee
        \be
        \texttt{integ~1:} \quad F^{c,a}+F^{c,b}-F^{a,c}-F^{b,c} = 0,
        \label{integ1_}
        \ee
        \begin{equation}
        \begin{gathered}
        \texttt{integ~2:} \quad F^{d,b}-F^{d,c}-F^{b,d}+F^{c,d}+F^{e,c}-F^{e,a}-F^{c,e}+F^{a,e}\\\qquad\qquad\qquad\qquad\qquad\,\,+F^{f,a}-F^{f,b}-F^{a,f}+F^{b,f}+2F^{c,b}-2F^{b,c} = 0.
        \label{integ2}
        \end{gathered}
        \end{equation}

   \item \textbf{Multiple Final Entries:} 
    Apart from the trivial zeros, there exist \textbf{26} final entry relations involving up to the last four letters.
    Here the specified adjacent slots can only be at the end of the key, and we highlight this difference by using $\mathcal{E}$ instead of $F$. When evaluated on the symbol, $\mathcal{E}$ is the same as $\mathcal{F}^{(L)}$.
        \be
        \texttt{final~0:} \quad \mathcal{E}^{a,d} = 0, \quad \quad \texttt{final~1:} \quad \mathcal{E}^{e,d} = 0, \nonumber
        \ee
        \be
        \texttt{final~2:} \quad \mathcal{E}^{a,d,d} = 0, \quad \quad \texttt{final~3:} \quad \mathcal{E}^{a,b,d} = 0, \nonumber
        \ee
        \be
        \texttt{final~4:} \quad \mathcal{E}^{a,c,e} = 0, \quad \quad \texttt{final~5:} \quad \mathcal{E}^{e,b,d} = 0, \quad \quad \texttt{final~6:} \quad \mathcal{E}^{e,d,d} = 0, \nonumber
        \ee
        \be
        \texttt{final~7:} \quad \mathcal{E}^{a,d,d,d} = 0, \quad \quad \texttt{final~8:} \quad \mathcal{E}^{a,b,b,d} = 0, \quad \quad \texttt{final~9:} \quad \mathcal{E}^{a,d,b,d} = 0, \nonumber
        \ee
        \be
        \texttt{final~10:} \quad \mathcal{E}^{c,b,b,d} = 0, \quad \quad \texttt{final~11:} \quad \mathcal{E}^{e,b,b,d} = 0, \quad \quad \texttt{final~12:} \quad \mathcal{E}^{e,b,d,d} = 0, \nonumber 
        \ee
        \be
        \texttt{final~13:} \quad \mathcal{E}^{e,d,b,d} = 0, \quad \quad \texttt{final~14:} \quad \mathcal{E}^{e,d,d,d} = 0, \quad \quad \texttt{final~15:} \quad \mathcal{E}^{f,d,b,d} = 0, \nonumber 
        \ee
        \be
        \texttt{final~16:} \quad \mathcal{E}^{b,f}-\mathcal{E}^{b,d} = 0,
        \label{final19_}
        \ee
        \be
        \texttt{final~17:} \quad \mathcal{E}^{c,d,d}+\mathcal{E}^{c,e,e} = 0,
        \label{final20_}
        \ee
        \be
        \texttt{final~18:} \quad \mathcal{E}^{d,d,b,d}-\mathcal{E}^{d,b,d,d} = 0,
        \label{final21_}
        \ee
        \be
        \texttt{final~19:} \quad \mathcal{E}^{c,b,d,d}-\mathcal{E}^{c,d,b,d} = 0,
        \label{final22}
        \ee
        \be
        \texttt{final~20:} \quad \mathcal{E}^{f,b,d}-\mathcal{E}^{d,b,d}+\mathcal{E}^{b,d,d} = 0,
        \label{final23}
        \ee
        \begin{equation}
        \begin{gathered}
        \texttt{final~21:} \quad \mathcal{E}^{b,d,d,d}+\mathcal{E}^{f,a,f,f}-\mathcal{E}^{d,b,d,d}-\mathcal{E}^{e,a,f,f}\\+\mathcal{E}^{f,b,d,d}-\mathcal{E}^{a,e,e,e} = 0,
        \label{final24}
        \end{gathered}
        \end{equation}
        \begin{equation}
        \begin{gathered}
        \texttt{final~22:} \quad \mathcal{E}^{a,b,d,d}-\frac{1}{2}\mathcal{E}^{c,d,d,d}-\frac{1}{2}\mathcal{E}^{d,c,e,e}+\frac{1}{2}\mathcal{E}^{a,e,e,e}\\+\frac{1}{2}\mathcal{E}^{e,a,f,f}-\frac{1}{2}\mathcal{E}^{f,a,f,f}+\frac{1}{2}\mathcal{E}^{e,c,e,e} = 0,
        \label{final25}
        \end{gathered}
        \end{equation}
        \begin{equation}
        \begin{gathered}
        \texttt{final~23:} \quad \mathcal{E}^{c,b,d,d}-\frac{1}{2}\mathcal{E}^{b,f,f,f}+\frac{1}{2}\mathcal{E}^{d,c,e,e}-\frac{1}{2}\mathcal{E}^{e,c,e,e}\\+\frac{1}{2}\mathcal{E}^{c,d,d,d}+\frac{1}{2}\mathcal{E}^{d,b,d,d}-\frac{1}{2}\mathcal{E}^{f,b,d,d} = 0,
        \label{final26}
        \end{gathered}
        \end{equation}
        \begin{equation}
        \begin{gathered}
        \texttt{final~24:} \quad \mathcal{E}^{c,d,b,d}-\frac{1}{2}\mathcal{E}^{b,f,f,f}+\frac{1}{2}\mathcal{E}^{d,c,e,e}-\frac{1}{2}\mathcal{E}^{e,c,e,e}\\+\frac{1}{2}\mathcal{E}^{c,d,d,d}+\frac{1}{2}\mathcal{E}^{d,b,d,d}-\frac{1}{2}\mathcal{E}^{f,b,d,d} = 0,
        \label{final27}
        \end{gathered}
        \end{equation}
        \begin{equation}
        \begin{gathered}
        \texttt{final~25:} \quad \mathcal{E}^{f,b,b,d}-\mathcal{E}^{d,b,b,d}+\mathcal{E}^{b,b,d,d}-\frac{1}{2}\mathcal{E}^{f,a,f,f}+\frac{1}{2}\mathcal{E}^{d,b,d,d}\\-\frac{1}{2}\mathcal{E}^{f,b,d,d}+\frac{1}{2}\mathcal{E}^{e,a,f,f}+\frac{1}{2}\mathcal{E}^{a,e,e,e}-\frac{1}{2}\mathcal{E}^{b,f,f,f}= 0.
        \label{final28}
        \end{gathered}
        \end{equation}

    The first sixteen relations ($\texttt{final~0}$ to $\texttt{final~15}$) are all one-term relations, restricting which letters \emph{cannot} appear at the end of any key. They are trivial for the model to learn and can be predicted perfectly after just a few epochs. We therefore omit a detailed assessment of these one-term relations.
    \item \textbf{Dihedral Symmetry:}
    The cycle and flip symmetries can both be expressed as two-term Group 1 relations because they require two keys related by the given symmetry to have identical coefficients. In order to evaluate the cycle and flip relations independently of each other, we select only one two-term relation instance of either kind from each 6-term dihedral orbit. 
    
\end{itemize}

\begin{figure}
\centering
\setlength{\tabcolsep}{-3pt}
\begin{tabular}{ccc}
\includegraphics[width=55mm]{rel_acc_final16_6loop.pdf} &
\includegraphics[width=55mm]{rel_acc_final18_6loop.pdf} &
\includegraphics[width=55mm]{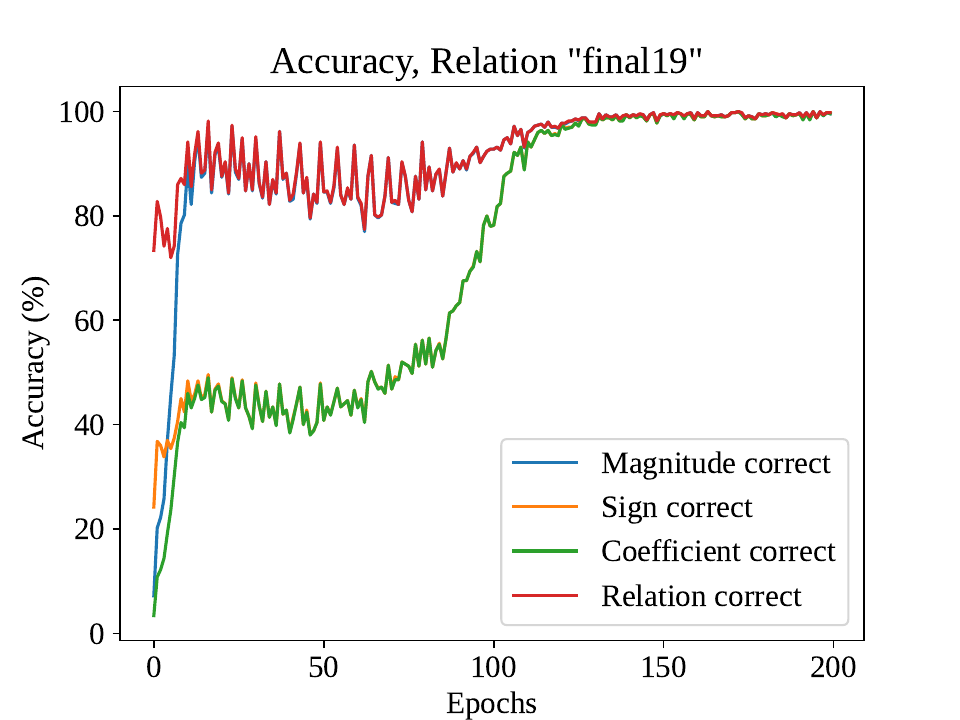} \\
\includegraphics[width=55mm]{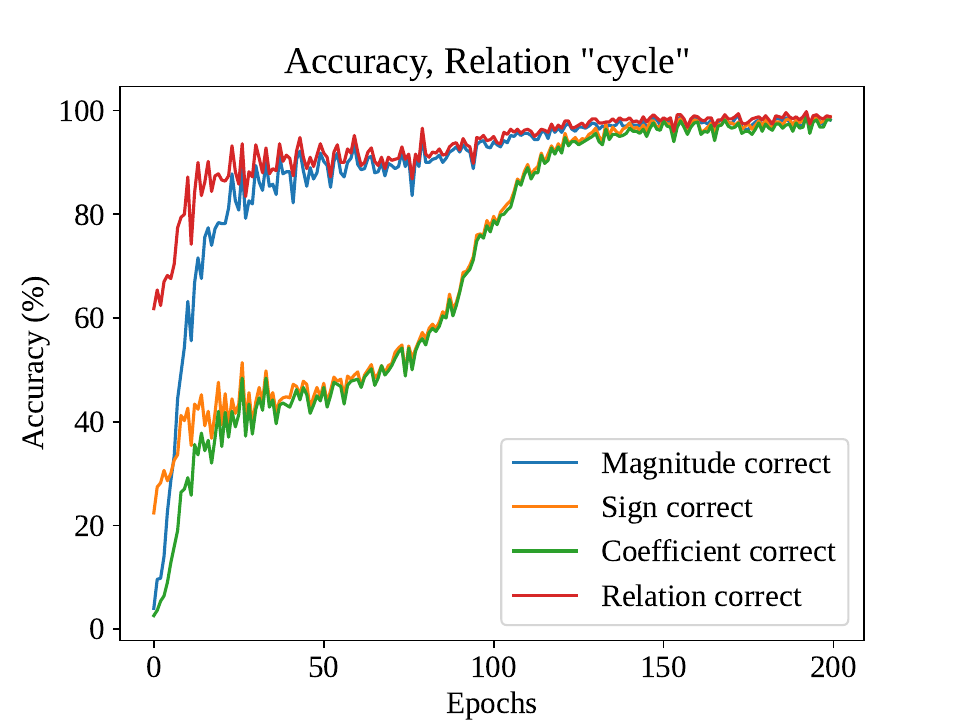} &
\includegraphics[width=55mm]{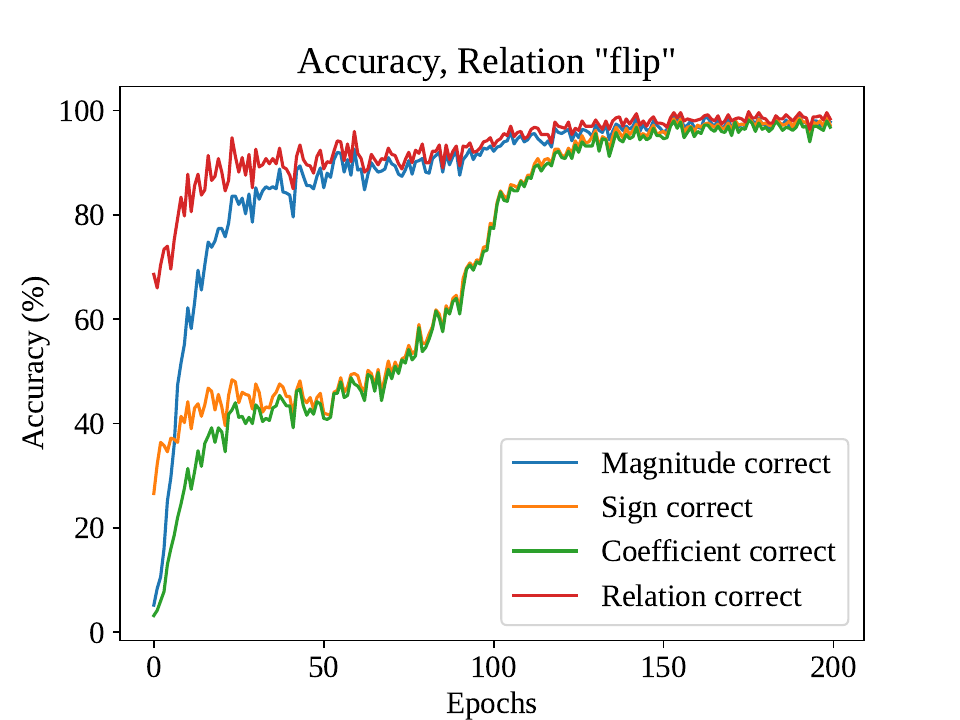}
\end{tabular}
\caption{Coefficient accuracy (green), magnitude accuracy (blue), relation accuracy (green), and sign accuracy (yellow) for all Group 1 relations.}
\label{fig:group1}
\end{figure}

\begin{figure}[htb]
\centering
\setlength{\tabcolsep}{-3pt}
\begin{tabular}{cc}
\includegraphics[width=55mm]{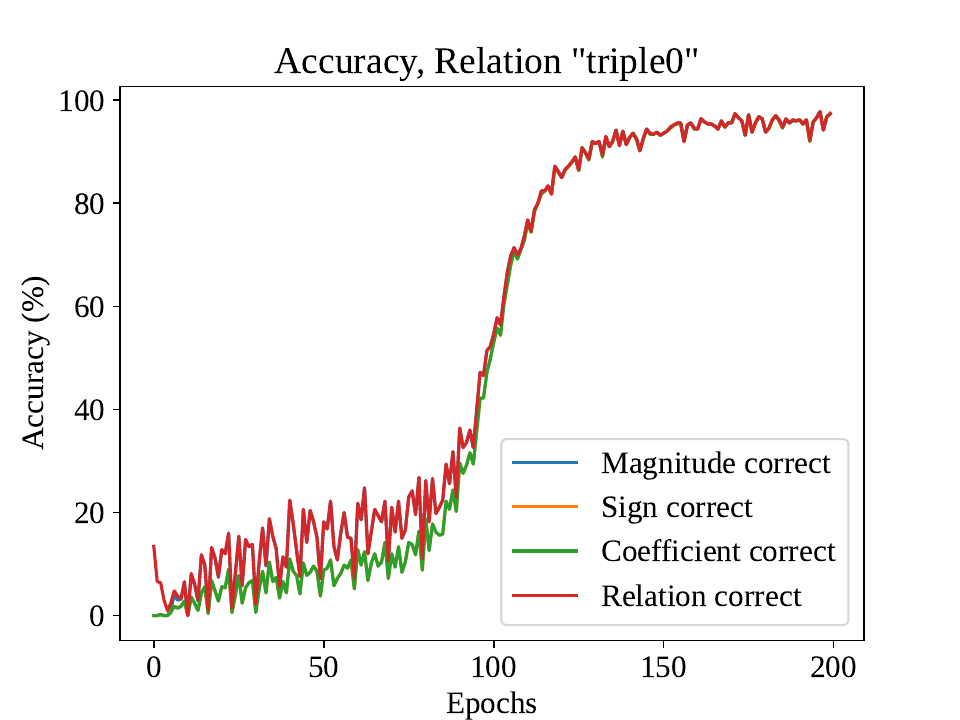} & \includegraphics[width=55mm]{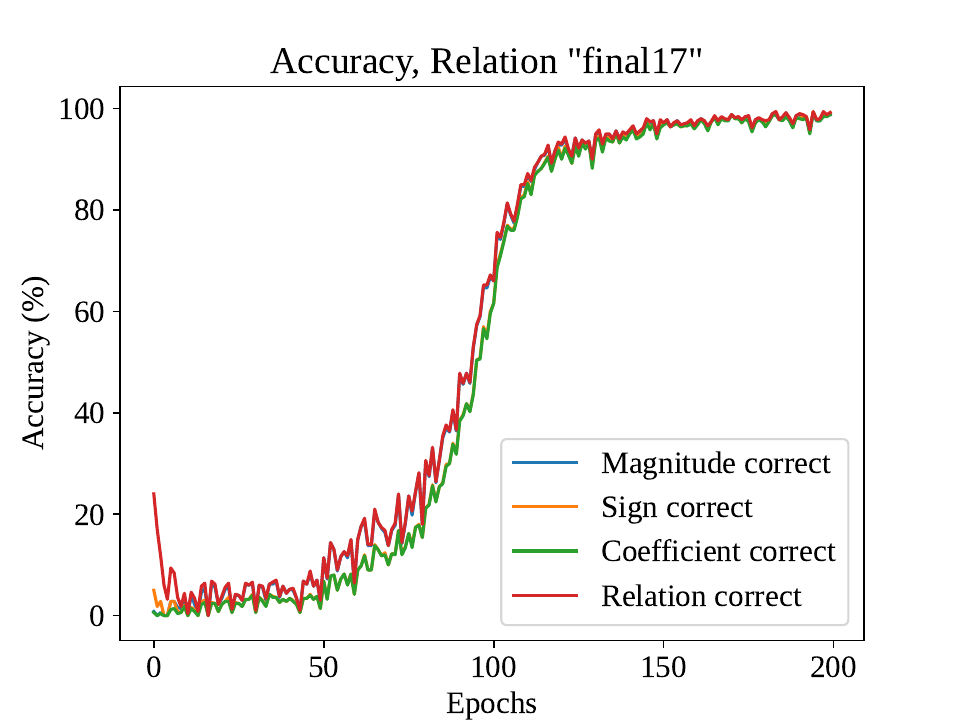} \\ 
\end{tabular}
\caption{Coefficient accuracy (green), magnitude accuracy (blue), relation accuracy (green), and sign accuracy (yellow) for all Group 2 relations.}
\label{fig:group2}
\end{figure}

\begin{figure}[htb]
\centering
\setlength{\tabcolsep}{-3pt}
\begin{tabular}{ccc}
\includegraphics[width=55mm]{rel_acc_integ0_6loop.pdf} & 
\includegraphics[width=55mm]{rel_acc_integ1_6loop.pdf} &
\includegraphics[width=55mm]{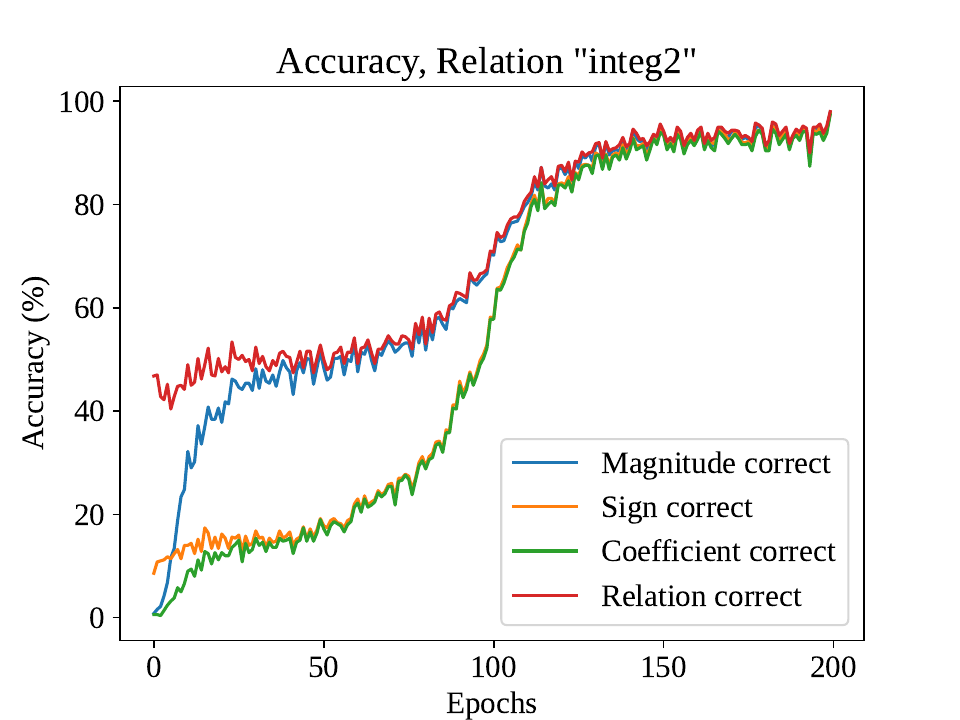} \\
\includegraphics[width=55mm]{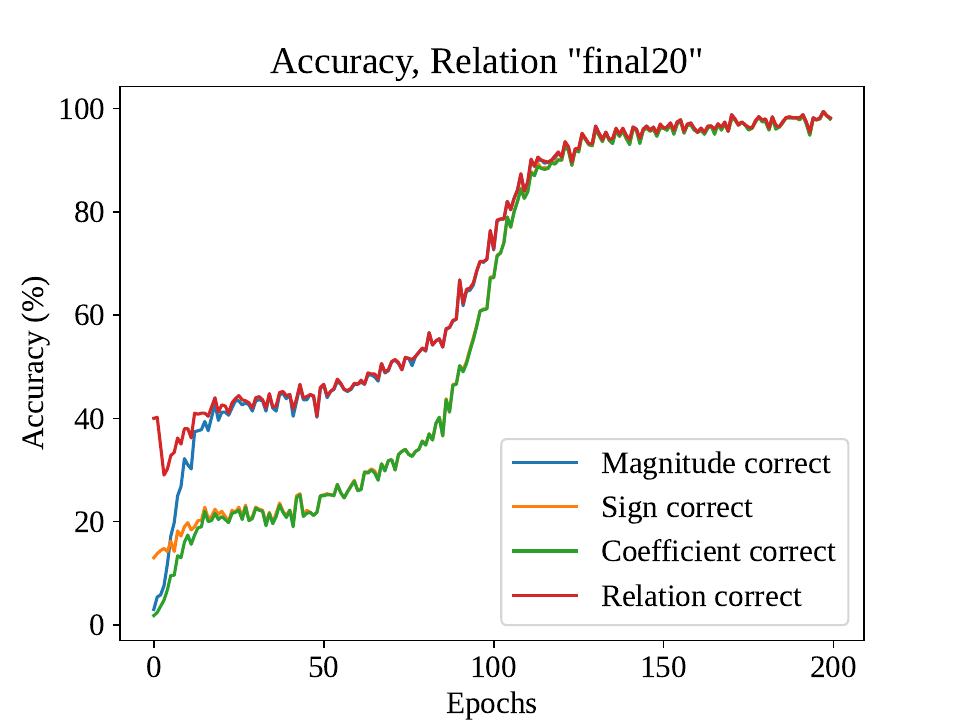} &
\includegraphics[width=55mm]{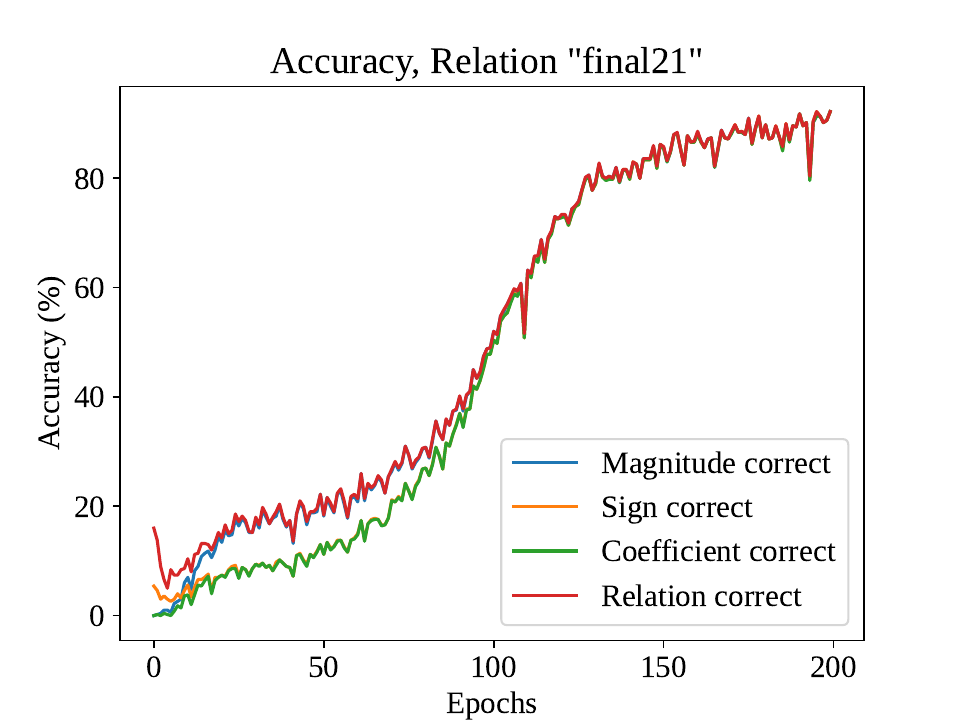} &
\includegraphics[width=55mm]{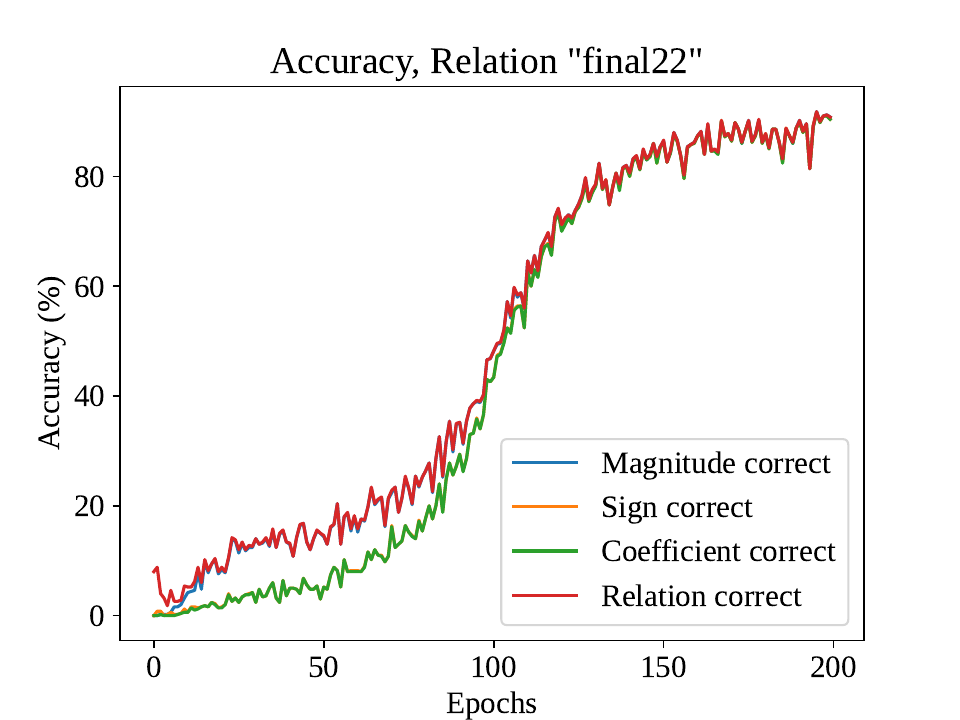} \\
\includegraphics[width=55mm]{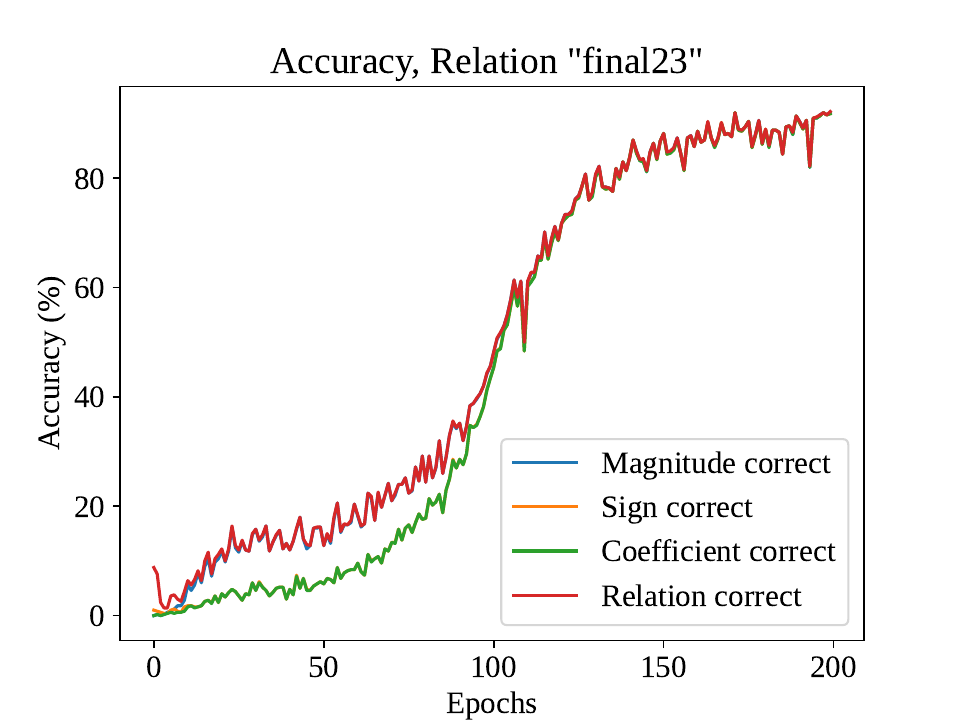} &
\includegraphics[width=55mm]{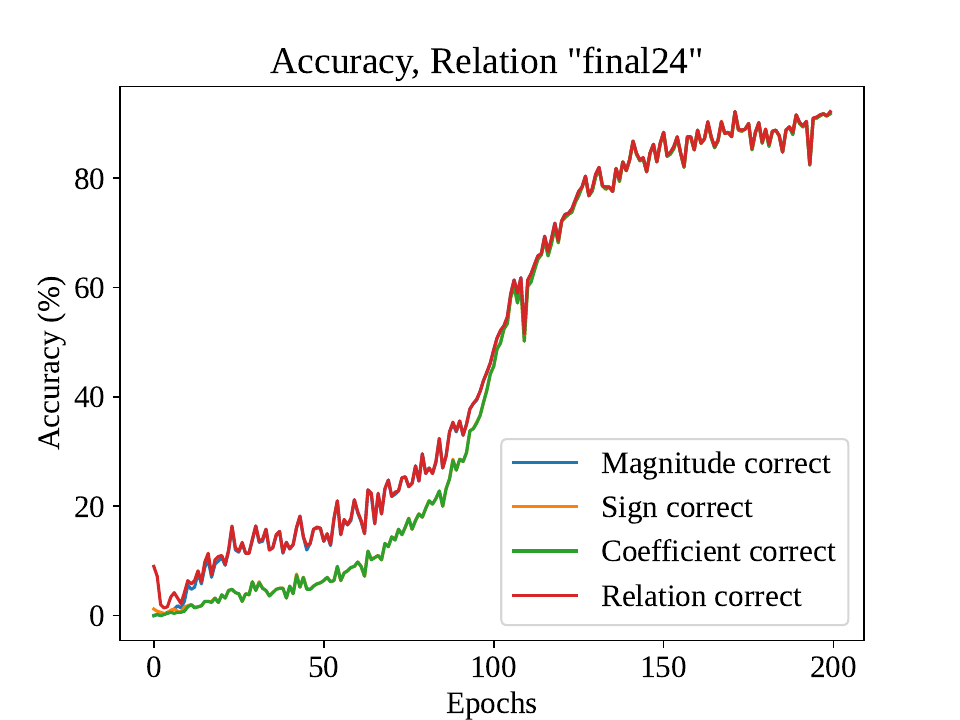} &
\includegraphics[width=55mm]{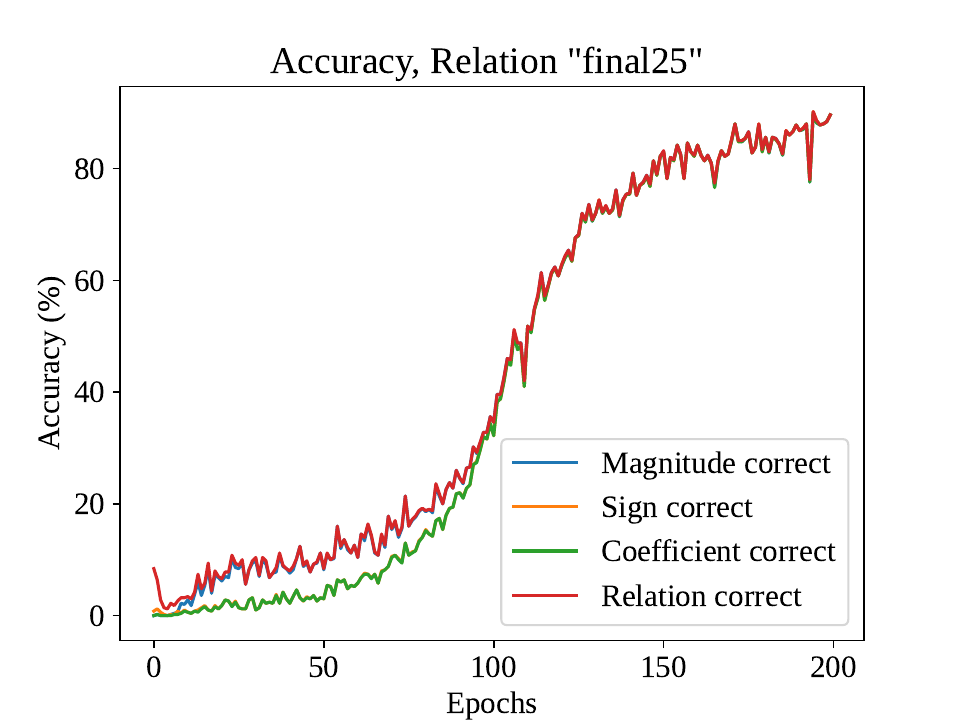} \\
\end{tabular}
\caption{Coefficient accuracy (green), magnitude accuracy (blue), relation accuracy (green), and sign accuracy (yellow) for all Group 3 relations.}
\label{fig:group3}
\end{figure}

We evaluate the models on the list of relations above using the same architecture, dataset, and accuracy criteria as in Section \ref{sec:relations}. We again categorize the relations according to the three groups mentioned in the main text. We plot the Group 1 results in Fig.~\ref{fig:group1}, the Group 2 results in Fig.~\ref{fig:group2}, and the Group 3 results in Fig.~\ref{fig:group3}. We note that Group 3 relations behave somewhat differently, depending on whether they are short (less than 4 terms) or long (more than 6 terms).

\section{Coefficient-from-Key: Model Ablations}
\label{app:ablations}

We perform a number of ablations in order to better understand the model performance for the coefficient-from-key experiment in Section~\ref{sec:coefffromword}, with an eye towards building a future model that can generalize across loops. Here we consider several different positional encoding schemes, explore encoder-only and decoder-only architectures, and vary both the number of layers in the model and the dimension $d$ of the encoded tokens. See Tables \ref{tab:arc_mods} and \ref{tab:arc_scan} for an overview of the results.

All models exhibit the two-phase learning structure, suggesting that this behavior is an intrinsic property of the problem itself rather than an artifact of the chosen model architecture. We report the epoch at which the test-set magnitude accuracy first exceeds $95\%$, the midpoint epoch of the plateau step (defined as the epoch at which the model's overall accuracy first reaches the midpoint between its final-state accuracy and $75\%$), the epoch at which the overall test-set accuracy first exceeds $95\%$, and the best overall test-set accuracy after 200 epochs.

\subsection{Architecture Ablations: Relative Position Encoding, Encoder-only, Decoder-only, and Non-Transformer Architectures}
\label{app:archs}

First, we test whether results change appreciably when the sign token is positioned at the end of the coefficient rather than the beginning, leaving all other hyperparameters unchanged from the baseline configuration. We next explore three variants of relative position encoding in the encoder of the baseline model of Section~\ref{sec:coefffromword}: (1) one in which learned relative position encoding is added to the product of the key and query matrices~\cite{Shaw} and implemented as in the TransformerXL~\cite{TransformerXL} and Conformer~\cite{Conformer} family of models; and (2-3) two rotary encodings---one standard (Rotary Position Embedding, a.k.a. RoPE~\cite{RoFormer}) and one long-context (xPos)~\cite{xPOS}. RoPE implements relative position encoding by multiplying the query and key matrices by a rotation matrix, while xPos introduces an additional scaling factor to enable better attention to tokens separated by large distances. Both are implemented in a common repository~\cite{lucidrains}. We compare to a baseline four-layer encoder-decoder Transformer with $d=256$. In all relative-position-encoding experiments, we only apply positional encoding in the encoder and do not use positional encoding at all in the autoregressive decoder (as in NoPE~\cite{NoPE}). For these experiments, we use the same zero percentage and nontrivial-zero biased prescription as in the model described in Section~\ref{sec:relations}. 

We find that models trained with relative position encoding slightly lag behind models trained with absolute position encoding, but achieve comparable test-set accuracy after the full 200 epochs. Additionally, we note that differences between RoPE and xPos are small.

Additionally, for the baseline model, varying the random seed three times can cause the point at which the test-set magnitude accuracy first exceeds 95\% (the first column of Table to shift by up to 5 epochs and the point at which the test-set magnitude accuracy first exceeds 95\% to shift by up to 8 epochs- thus, many of the differences in performance between architectures are within this window of uncertainty.

Our second class of ablations explores different Transformer architectures. In addition to the encoder-decoder Transformer used for the majority of our experiments, we train both an encoder-only Transformer and a decoder-only Transformer and evaluate their performance. The encoder-only (or decoder-only) model has $8$ layers in the encoder (or decoder), $d=256$, and $8$ attention heads. We also train an encoder-only Transformer with $8$ layers, $d=256$, and $8$ attention heads using the RoPE rotary positional embedding. 

In order to perform training in a decoder-only paradigm, we concatenate coefficient and key in a ``prompt-completion'' manner similar to that of the GPT family of models \cite{GPT}. When evaluating the predictions of the decoder-only model, we only evaluate accuracy on the ``completion'' portion, which is the tokens corresponding to the integer coefficients. Although the model is trained on next-token prediction on the full input, whether it correctly outputs the next letter in the key is not a well-defined objective, as many possible keys exist. A single solution exists only once the model has seen all tokens corresponding to the full key and is asked to predict the remaining tokens corresponding to the coefficient.

Likewise, when evaluating the predictions of the encoder-only model, all model outputs will be exactly $2L$ tokens long (the same length as the input). Because the coefficient sequence encoded in base-1000 encoding will be much shorter than $2L$, the encoder-only model will produce more output tokens than there are tokens in the coefficient sequence. We require that the tokens at the start of this output sequence correspond to the tokens of the coefficient sequence and that the output token immediately after the coefficient sequence corresponds to an end-of-sequence token. We ignore all tokens in the encoder output that appear after this end-of-sequence token.

\begin{table}[h!]
    \small
    \centering
    \begin{tabular}{lccccc}
        \toprule
        Architecture & \CellWithForcedBreak{First Mag. Acc. \\ > 95\% [Epoch]} & \CellWithForcedBreak{Midpoint of \\ Step [Epoch]} & \CellWithForcedBreak{First Total Acc. \\ > 95\% [Epoch]} & \CellWithForcedBreak{Best Acc. \\ Epoch 200} \\
        \midrule
        \textbf{$4/4$ enc-dec} & 26 & 94 & 105 & 99.2\% \\
        $4/4$ enc-dec, sign last & 27 & 102 & 114 & 99.5\% \\
        $4/4$ enc-dec, rel\_pos & 34 & 98 & 105 & 98.4\% \\
        $4/4$ enc-dec, RoPE & 33 & 121 & 132 & 99.1\% \\
        $4/4$ enc-dec, xPos  & 38 & 121 & 133 & 99.3\%\\
        $8$-layer, enc-only & 28 & 107 & 115 & 99.5\%\\
        $8$-layer, enc-only, RoPE & 36 & 122 & 131 & 99.4\%\\
        $8$-layer, dec-only & 39 & 105 & 112 & 99.2\% \\
        $8$-layer, dec-only, sign last & 39 & 118 & 134 & 99.3\% \\
       \bottomrule
    \end{tabular}
    \caption{Model learning dynamics for a variety of architectures trained and evaluated on the same $L=6$ dataset. All display the double-plateau structure. We indicate the epoch at which the test-set magnitude accuracy first exceeds 95\%, the midpoint of the plateau step, the epoch at which the overall test-set accuracy first exceeds 95\%, and the best overall test-set accuracy after 200 epochs. Numbers reported are for the best result of three seeds, with `best' defined as the first model to reach 95\% accuracy, or, if this threshold is not reached by any of the three, the model with the highest overall test-set accuracy after 200 epochs. The baseline model is indicated in bold.
    }
    \label{tab:arc_mods}
\end{table}

Finally, we perform the nonzero coefficient-from-key prediction task at 5 loops using non-Transformer architectures. Specifically, we use an encoder-decoder Long-Short-Term Memory recurrent neural network (LSTM) \cite{HochSchm97} and a Gated Recurrent Unit (GRU) \cite{ChoGRU}. The embedding dimension is $d=512$ and the number of layers in the encoder and decoder are both taken to be 2, as in the Transformer, while the hidden layer size is either chosen such that the number of trainable parameters is approximately the same as in the Transformer, or chosen to be $h=2048$. We again observe the familiar two-phase behavior as in Fig.~\ref{fig:loop56}. We report the results compared to a Transformer model in the table~\ref{tabLSTM}. For the models with a comparable number of parameters as the Transformer, we observe that performance is slightly worse. However, when $h=2048$, the LSTM performs at least as well as the Transformer on this particular task, and the GRU is not qualitatively worse.

\begin{table}[h!]
    \small
    \centering
    \begin{tabular}{lccccc}
        \toprule
        Architecture & \CellWithForcedBreak{First Mag. Acc. \\ > 95\% [Epoch]} & \CellWithForcedBreak{Midpoint of \\ Step [Epoch]} & \CellWithForcedBreak{First Total Acc. \\ > 95\% [Epoch]} & \CellWithForcedBreak{Best Acc. \\ Epoch 15} & $N_{\text{params}}$\\
        \midrule
        \textbf{Transformer} & 3 & 6 & 9 & 99.10\% & 20M \\
        LSTM ($h=786$) & 7 & 12 & N/A & 87.42\% & 20M\\
        GRU ($h=916$) & 5 & 10 & N/A & 90.91\% & 20M\\
        GRU ($h=2048$) & 3 & 5 & 6 & 98.88\% & 85M\\
        LSTM ($h=2048$) & 3 & 5 & 6 & 99.76\% & 112M\\
       \bottomrule
    \end{tabular}
    \caption{Model learning dynamics for a variety of LSTM architectures trained and evaluated on the same $L=5$ dataset. All display the double-plateau structure. We indicate the epoch at which the test-set magnitude accuracy first exceeds 95\%, the midpoint of the plateau step, the epoch at which the overall test-set accuracy first exceeds 95\%, and the best overall test-set accuracy after 15 epochs. The baseline Transformer model is indicated in bold. For this experiment, we use only one seed rather than the best of 4.}
    \label{tabLSTM}
\end{table}

%%%%%%%%%%%%%%%%%%%%%%%%%%%%%%%%

\subsection{Architecture Parameter Scan}
\label{app:archscan}

We now scan a variety of model depths and sizes for the $L=6$ coefficient-from-key prediction task, with results given in Table \ref{tab:arc_scan}. We vary the number of layers in both the encoder and decoder and the internal dimension $d$ of the Transformer. For each model, we report the number of epochs required to reach a variety of benchmarks that summarize the two-phase learning behavior: (1) the number of epochs required to reach the first phase plateau ($>95\%$ magnitude accuracy); (2) the number of epochs required to reach the midpoint of the step between the two plateaus (i.e., the midpoint between $75\%$ overall accuracy and the value of overall accuracy after 200 epochs); and (3) the number of epochs required to reach $95\%$ overall accuracy. We also report the best overall accuracy after 200 epochs.

Qualitatively, while all models exhibit the two-phase learning behavior, both phases are compressed (last for fewer epochs) as model capacity increases. The relationship between the number of training epochs required to reach a particular benchmark and the model dimension appears to be approximately linear. That is, increasing model dimension by a factor of two while leaving all other hyperparameters fixed will cause benchmarks to be reachable in half as many epochs. The number of attention heads does not appear to have as much impact on learning dynamics as the model dimension and depth.

\begin{table}[h]
    \small
    \centering
    \begin{tabular}{lccccc}
        \toprule
        Architecture & \CellWithForcedBreak{First Mag. Acc. \\ > 95\% [Epoch]} & \CellWithForcedBreak{Midpoint of \\ Step [Epoch]} & \CellWithForcedBreak{First Total Acc. \\ > 95\% [Epoch]} & \CellWithForcedBreak{Best Acc. \\ Epoch 200} \\
        \midrule
        \textbf{$4/4, d=256, h = 8$} & 26 & 94 & 105 & 99.2\% \\
        $4/4, d=256, h = 4$  & 29 & 100 & 108 & 99.1\% \\ 
        $4/4, d=512, h = 8$ & 15 & 54 & 62 & 99.4\% \\
        $4/4, d=512, h = 4$ & 19 & 65 & 75 & 99.3\% \\
        $2/2, d=256, h = 8$  & 45 & 186 & N/A & 91.7\% \\
        $2/2, d=256, h = 4$  & 63 & N/A & N/A & 75.5\% \\
        $2/2, d=512, h = 8$ & 26 & 93 & 107 & 98.9\% \\
        $2/2, d=512, h = 4$  & 29 & 120 & 133 & 98.3\% \\
       \bottomrule
    \end{tabular}
    \caption{Model learning dynamics for a variety of model sizes trained and evaluated on the same $L=6$ dataset.
    All display the double-plateau structure. We indicate the epoch at which the test-set magnitude accuracy first exceeds 95\%, the midpoint of the plateau step, the epoch at which the overall test-set accuracy first exceeds 95\%, and the best overall test-set accuracy after 200 epochs. Numbers reported are for the best result of three seeds, with `best' defined as the first model to reach 95\% accuracy, or, if this threshold is not reached by any of the three, the model with the highest overall test-set accuracy after 200 epochs. The baseline model is indicated in bold.
    }
    \label{tab:arc_scan}
\end{table}

Additionally, because we only split our data into training and test sets, rather than training, validation, and test sets, the accuracy metrics quoted may be unduly influenced by the specific choice of test set. In order to demonstrate that performance does not degrade appreciably for different choices of test set, we perform the 5-loop nonzero coefficient-from-key prediction task with three different random training/test set configurations, but the same architecture and hyperparameter configuration as in the baseline experiment in Section \ref{sec:coefffromword}. We report these results in Table \ref{tab:traintest}. The accuracies for the different splits are a little lower than for the baseline.

\begin{table}[h!]
    \small
    \centering
    \begin{tabular}{lccccc}
        \toprule
        Architecture & \CellWithForcedBreak{First Mag. Acc. \\ > 95\% [Epoch]} & \CellWithForcedBreak{Midpoint of \\ Step [Epoch]} & \CellWithForcedBreak{First Total Acc. \\ > 95\% [Epoch]} & \CellWithForcedBreak{Best Acc. \\ Epoch 15} \\
        \midrule
        Baseline & 3 & 6 & 9 & 99.10\% \\  
        Split 1 & 2 & 6 & 9 & 98.61\% \\
        Split 2 & 3 & 7 & 9 & 98.40\% \\
        Split 3 & 3 & 7 & 11 & 98.92\% \\
       \bottomrule
    \end{tabular}
    \caption{Model learning dynamics for four different choices of random training and test set split on the 5-loop coefficient-from-key task. We indicate the epoch at which the test-set magnitude accuracy first exceeds 95\%, the midpoint of the plateau step, the epoch at which the overall test-set accuracy first exceeds 95\%, and the best overall test-set accuracy after 15 epochs.}
    \label{tab:traintest}
\end{table}

%%%%%%%%%%%%%%%%%%%%%%%%%%%%%%%%%%%%%%%%%%%%%%%%%%%%%%%%%%%%%%%%%%%%%%%%%%%%%%%
\section{Further Strike-out Experiments}\label{app:recurrence}

Table~\ref{tab:ablation_recurrence} presents the overall, magnitude and sign accuracy of models trained for up to $700$ epochs, for different variations of the strike-two method, performed on loop $L=6$ data as in the main text. 
We use $4$-layer Transformers with dimension $d=512$ and $8$ attention heads. 
All models are trained for up to 700 epochs, and we indicate the epoch at which the indicated best accuracy is first achieved. The experiments for the ``strike-two parents'',  ``shuffled parents'', ``parent signs only'' and ``parent magnitudes only'' sections of Table~\ref{tab:ablation_recurrence} are described in Section \ref{sec:strikeout}; Table~\ref{tab:ablation_recurrence} provides more information about their dependence on the strike-out distance $k$.

Here, two new sets of experiments are introduced. In the ``sorted unique'' experiments, all parents are sorted in ascending order, and duplicate parent coefficients are removed; thus, the multiplicity of each of the parent coefficients, a permutation-invariant property, is eliminated. This model proves harder to train, with the best models achieving an accuracy of $83.4\%$ after $350$ epochs. Still, it is much better than random guessing, with sign information almost totally recoverable ($93.6\%$). This result suggests that some amount of information useful to reconstructing the coefficient is present even when multiplicities are removed.

In the ``zero/nonzero'' experiments, all nonzero parent coefficients are encoded as ``$1$'', while all zero parent coefficients are encoded as “0”. Such experiments provide further information on the ``signs-only'' experiment in the main text. While clarifying which parents are positive, negative, and zero is enough to reconstruct the sign and magnitude of the target coefficient at $99.0\%$ and $93.5\%$ respectively, providing only which parents are zero or nonzero causes magnitude accuracy to drop to $60.1\%$ and sign accuracy to drop to $61.7\%$. Thus, zero/nonzero information alone is not enough to reconstruct the target coefficient.

\begin{table}[h]
    \small
    \centering
    \begin{tabular}{lcccccc}
        \toprule
          & Distance & \multicolumn{1}{p{1.25cm}}{\centering Best \\ Epoch} & \multicolumn{1}{p{1.25cm}}{\centering Train \\ Size} &  Accuracy & \multicolumn{1}{p{1.25cm}}{\centering Magnitude \\ Accuracy} & \multicolumn{1}{p{1.25cm}}{\centering Sign \\ Accuracy}\\
        \midrule
        Strike-two parents & Full & 455 & 757,500 & 98.1\% & 98.4\% & 99.6\% \\
         & 5 & 524 & 754,060 & 98.3\% & 98.6\% & 99.7\% \\
         & 3 & 601 & 738,352 & 98.4\% & 98.7\% & 99.7\% \\
         & 2 & 681 & 688,869 & 98.1\% & 98.3\% & 99.5\% \\
         & 1 & 646 & 576,510 & 94.3\% & 95.2\% & 98.5\% \\
        \midrule
        
        Shuffled parents & Full & 407&  4,906,466 & 95.2\% & 99.1\% & 96.3\% \\
         & 5 & 376& 4,906,466 & 94.7\% & 98.8\% & 95.8\% \\
         & 3 & 408 & 4,906,436 & 95.1\% & 99.0\% & 96.1\% \\
         & 2 & 433 & 4,906,249 & 93.5\% & 98.1\% & 95.0\% \\
         & 1 & 442 & 4,882,501 & 91.1\% & 92.2\% & 96.5\% \\
        \midrule

        Sorted parents & Full & 389 & 591,864 & 91.5\% & 93.8\% & 96.6\% \\
         & 5 &  432 & 717,534 & 93.9\% & 95.4\% & 97.9\% \\
         & 3 & 514 & 702,363 & 93.1\% & 94.6\% & 97.5\% \\
         & 2 & 453 & 657,863 & 90.7\% & 92.4\% & 96.5\% \\
         & 1 & 459 & 536,588 & 76.8\% & 79.3\% & 90.5\% \\
        \midrule

        Sorted unique parents & Full & 329 & 476,932 & 79.0\% & 84.0\% & 92.1\% \\
        & 5 & 355 & 538,325 & 83.4\% & 87.4\% & 93.6\% \\
        & 3 & 349 & 487,813 & 79.7\% & 84.1\% & 91.7\% \\
        & 2 & 287 & 436,012 & 73.6\% & 78.5\% & 89.4\% \\
        & 1 & 314 & 355,147 & 57.2\% & 61.9\% & 80.9\% \\

        \midrule
        Zero / nonzero & Full & 304 & 497,112 & 40.7\% & 60.1\% & 61.7\% \\ 
        & 5 & 229 & 467,871 & 35.0\% & 53.8\% & 59.4\% \\ 
        & 3 &  93 & 415,230 & 22.9\% & 39.1\% & 54.9\% \\ 
        & 2 &  18 & 344,831 & 10.1\% & 20.1\% & 50.3\% \\ 
        & 1 &   1 & 131,812 &  0.8\% & 1.4\% & 50.4\% \\ 
        \midrule
        
        Parent signs only & Full & 404 & 748,088 & 93.3\% & 93.5\% & 99.0\% \\
        & 5 &  330 & 739,479 & 92.4\% & 92.5\% & 99.0\% \\
        & 3 &  294 & 711,450 & 88.2\% & 88.4\% & 98.2\% \\
        & 2 &  331 & 653,368 & 73.3\% & 73.6\% & 94.8\% \\
        & 1 &  22 & 468,339 &  5.8\% & 6.5\% & 64.6\% \\
        \midrule

        Parent magnitudes only & Full & 395 & 751,675 & 81.8\% & 98.4\% & 83.2\% \\
        & 5 & 445 & 747,187 & 81.2\% & 98.4\% & 82.4\% \\
        & 3 & 452 & 726,554 & 79.7\% & 98.3\% & 80.9\% \\
        & 2 & 611 & 672,561 & 77.8\% & 97.9\% & 79.4\% \\
        & 1 & 509 & 527,843 & 64.1\% & 93.6\% & 67.7\% \\

        \bottomrule
    \end{tabular}
    \small
    \caption{\small \textbf{Overall, magnitude and sign accuracy for strike-two method.} We show the best of four seeds, and for a variety of experiments and $k$-values we quote the epoch at which the best overall accuracy was attained, the training set size after duplicate removal, and the best values of overall, magnitude and sign accuracy after 700 epochs.}
    \label{tab:ablation_recurrence}
    \end{table}

\end{document}